%% file: arxiv.tex
\documentclass[runningheads]{llncs}

\usepackage{subfig}
\usepackage{url}
\usepackage{nicefrac}
\usepackage{tabularx}
\usepackage{afterpage}

\usepackage{rotating}
\usepackage{fancyhdr}  

\usepackage{color}
\usepackage{mathtools}
\usepackage{cite}
\usepackage{gensymb}
\usepackage{amsfonts}
\usepackage{amssymb}
\usepackage{amsmath}
\usepackage{graphicx,wrapfig}
\usepackage{caption}
\usepackage[lined, boxed]{algorithm2e}

\usepackage[inline]{enumitem}
\usepackage{nicefrac}
\usepackage{hyperref}
\usepackage{tikz}
\usepackage{tabularx, rotating, caption}
\usepackage{booktabs}

\usepackage{pgfplots}
\pgfplotsset{compat=newest}

\usepackage{filecontents} 
\usepackage{pgfplots}
\usepackage{pgfplotstable}
\usepackage{blindtext}
\usepackage{siunitx}
\usepackage{tikz} 

\usepackage{pgfplotstable}
\usetikzlibrary{spy,shadows}
\tikzset{spy using overlaysshadow/.style={
    spy scope={#1,
         every spy on node/.style={
            circle,
            fill, fill opacity=0.2, text opacity=1
             },
         every spy in node/.style={
                 circle, circular drop shadow,
                 fill=white, draw, ultra thick, cap=round
            }
        }
    }
}

\usepgfplotslibrary{units, fillbetween} 
\usetikzlibrary{patterns}
\usetikzlibrary{external}

\DeclareMathOperator*{\argmax}{argmax}

\title{\Large \bf 
Exploration of Indoor Environments Predicting\\ the Layout of Partially Observed Rooms}

\author{Matteo Luperto\inst{1} \and Luca Fochetta\inst{2} \and Francesco Amigoni\inst{2}}
\institute{Universit\`{a} degli Studi di Milano, 20122 Milano, Italy\\
\email{matteo.luperto@unimi.it}
\and Politecnico di Milano, 20133 Milano, Italy,\\
\email{luca.fochetta@mail.polimi.it}, \email{francesco.amigoni@polimi.it}}

\begin{document}

\def\datafile{./data/allData.csv}

\maketitle
\thispagestyle{empty}
\pagestyle{empty}

\begin{abstract}

We consider exploration tasks in which an autonomous mobile robot incrementally builds maps of initially unknown indoor environments. In such tasks, the robot makes a sequence of decisions on where to move next that, usually, are based on knowledge about the observed parts of the environment. 
In this paper, we present an approach that exploits a prediction of the geometric structure of the unknown parts of an environment to improve exploration performance. In particular, we leverage an existing method that reconstructs the layout of an environment starting from a partial grid map and that predicts the shape of partially observed rooms on the basis of geometric features representing the regularities of the indoor environment. Then, we originally employ the predicted layout to estimate the amount of new area the robot would observe from candidate locations in order to inform the selection of the next best location and to early stop the exploration when no further relevant area is expected to be discovered. Experimental activities show that our approach is able to effectively predict the layout of partially observed rooms and to use such knowledge to speed up the exploration.

\end{abstract}

\input{01-Intro}
\input{02-Art}

\input{03-Our}

\input{04-Exp}
\input{05-Con}


\bibliographystyle{splncs04}
\bibliography{citations}



\end{document}

%% file: 01-Intro.tex
\section{Introduction}

For autonomous mobile robots, \emph{exploration} is a task that incrementally builds maps of initially unknown environments~\cite{Thrun02roboticmapping:}. Typically, at each stage of the exploration process, a robot selects the next best location (often on a frontier between known and unknown space in the current map) according to an \emph{exploration strategy}~\cite{doi:10.1177/0278364902021010834}. The robot iteratively reaches the selected location, acquires new knowledge on the environment, updates the map, and selects a new next best location, until the environment is fully observed. Decisions made by most exploration strategies are only informed by the knowledge of the observed part of the environment that is represented by the current map~\cite{doi:10.1177/0278364902021010834,Basilico2011}. However, structured indoor environments show regularities and symmetries~\cite{luperto2013} that could be exploited to better inform the selection of the next best locations.  

In this paper, we present a method that exploits the prediction of the geometric structure of the unknown part of an indoor environment to select the next best location for an exploring mobile robot and to terminate the exploration early if no further relevant area is expected to be observed. 

More precisely, we consider a mobile robot that explores an initially unknown indoor environment in order to build its 2D grid map (Section~\ref{sec:EXPL}). At each stage of the exploration process, we reconstruct, from the current grid map, the layout of the observed part and we predict the layout of the unobserved part of the environment. The \emph{layout} is an abstract geometrical representation in which rooms are modeled as polygons, capturing their shape and filtering out noisy data (e.g., misalignment of walls and small pieces of furniture) that are present in grid maps~\cite{liu2014generalizable,ARMENI,mura2014automatic,IAS15}. For layout reconstruction, we employ a method we previously developed~\cite{IAS15,ICRA19} (summarized in Section~\ref{sec:LAY}). The shape of partially observed rooms is predicted following the insight that different parts of the building share common features. For example, rooms connected to the same corridor likely share a common wall and have similar shapes, as it usually happens in large buildings like schools and offices \cite{neufert2012architects}.
We originally exploit the predicted layout to evaluate the amount of new area that the robot expects to perceive from the candidate locations on current frontiers, to inform its decision on where to go next (Section~\ref{S:SOL}). Moreover, if the amount of the unobserved area returned by the predicted layout is below a threshold, the exploration is stopped. 

Experimental activities, conducted in several simulated large-scale indoor environments, show that our method is able to effectively exploit the predicted layout of partially observed rooms to speed up the exploration in a wide range of situations (Section~\ref{sec:EXP}). In general, the more the exploration progresses, the more knowledge about the structure of the environment the robot acquires, the more the layout prediction becomes accurate, and the more gain our proposed method provides wrt classical exploration strategies that only consider knowledge about the observed part of the environment that is contained in the grid map.

The main original contribution of this paper is thus a method that employs the predicted layout of a partially observed environment to speed up the exploration. The contribution of this paper is different from that of~\cite{AAMAS19,ecmr2019}, where we assume to know the floor plans of indoor environments in advance, before starting the exploration. In this paper, instead, we retrieve a layout that predicts the shape of partially observed rooms at each stage during the exploration process.

%% file: 02-Art.tex
\section{Related Work}
\label{S:relatedwork}
Exploration is the incremental process with which a robot (or a multirobot system) covers with its sensors an initially unknown environment. 
Two main families of approaches have been developed for exploration: \emph{frontier-based}, which move the robots to the geometrical boundaries between known and unknown portions of environments~\cite{613851}, and \emph{information-based}, which move the robots to the most informative locations, according to some information measure (e.g.,~\cite{Stachniss2005InformationGE}). In this paper, we focus on the first family of approaches, since they more naturally address the discovery of space for the task of map building we consider.

Different exploration strategies have been proposed to select the next best frontier, all of them being greedy~\cite{1249657}, due to the inherently on-line nature of the exploration problem. A complete survey is out of the scope of this paper (the reader can refer, e.g., to~\cite{julia-exploration-survey}), so we just report some examples of exploration strategies. For instance,~\cite{doi:10.1177/0278364902021010834} evaluates each candidate frontier taking into account its distance from the robot's current position and the expected information gain (in terms of the maximum unexplored area that could be viewed from it). The two criteria are combined in an exponential utility function. Also~\cite{Tovar2006} combines criteria related to distance and information gain in an \emph{ad hoc} utility function.
In~\cite{Basilico2011} and~\cite{1570708}, more principled ways to aggregate criteria, based on multi-objective optimization, are proposed. In all the above cases, exploration strategies choose the next best frontier according to the knowledge on the part of the environment that has been already observed, not considering what the robot has not yet observed. 

The use of other forms of knowledge to integrate the information from the current map has been investigated with the aim of improving the performance of exploration. 
In \cite{PereaStrom2017125}, the possible aspect of the unexplored part of an environment is predicted by exploiting a database of previously mapped environments, in order to complete the partial map obtained by the robot.
A similar approach, but extended to multirobot settings, is that of \cite{Smith2018}.
Unlike our approach, both \cite{PereaStrom2017125} and \cite{Smith2018} use knowledge relative to environments different from the one where the robot operates. Hence, while the above methods rely on the presence of libraries of environments observed in the past, our approach can be applied also when such data are not available.

The authors of \cite{7390002} propose an exploration approach that, knowing a representation of the environment in terms of a topo-metric graph, finds an efficient exploration path. 
Similarly to ours, this method exploits the knowledge of the same environment in which the robot operates. However, in~\cite{7390002} (and in~\cite{AAMAS19,ecmr2019} mentioned before), the robot is provided with prior knowledge about the environment, while in our approach the robot updates and exploits the knowledge as the exploration progresses. 

A method that shares some similarities with our approach is that of~\cite{p-slam}, which predicts the structure of an unexplored region of an environment to improve SLAM performance. The method tries to reconstruct the neighborhood of a frontier by identifying similar structures in the known map. 
The prediction of \cite{p-slam} considers the local similarity between different parts of the same environment, while our approach considers more abstract global features like the fact that rooms aligned along the same corridor share the same wall. 

Another recent method that is similar to ours is that of~\cite{LearnedMap}, where a variational autoencoder (VAE) is employed to predict unknown regions of an environment starting from a partial map. The prediction is then used to compute the expected information gain for candidate locations. The good performance of this approach is related to the fact that the authors consider buildings that are very similar to each other (see data and discussion in~\cite{lupertoIAS13}). The generalization of the approach to other environments requires to use more diversified data. Moreover, the method in  \cite{LearnedMap} considers empty maps (a VAE could be trained with real-world maps with clutter and furniture, but this would require a significantly larger training set). Our method, instead, can be used with any map (in which the walls can be identified) acquired in any environment.

We finally mention some methods that infer the presence and location of specific elements in the unknown part of environments, like emergency exits~\cite{CaleyHollinger}, labels of unseen rooms~\cite{pronobis2012large}, and portions of environments represented as graphs~\cite{aydemir2012,luperto2018predicting}. These methods do not provide the geometrical information we exploit for speeding up exploration.

%% file: 03-Our.tex
\section{Our Approach}\label{sec:OUR}

In this section, we describe the exploration process we consider (Section~\ref{sec:EXPL}), then we detail the methods we use for reconstructing and predicting the layout starting from a partial grid map (Section~\ref{sec:LAY}) and for exploiting the predicted layout to estimate the amount of information the robot can acquire from a frontier (Section~\ref{S:SOL}). Finally, we illustrate the use of the predicted layout to implement an early stopping criterion for the exploration (Section~\ref{s:es}).

\subsection{Overview of the exploration process}\label{sec:EXPL}

We consider a single mobile robot that explores an initially unknown planar indoor environment $E$ using a 2D laser range scanner with a given field of view and range. 
The exploration process is frontier-based and is composed of the following steps (details are provided below):
\begin{enumerate}
\item[(a)] the robot perceives a portion of $E$ from its current location $p_{R}$ using the laser range scanner and integrates the new perception in the current map $M$,
\item[(b)] the robot identifies the current frontiers on $M$, namely the boundaries between known and unknown space, and extracts from them the set $C$ of candidate locations,
\item[(c)] the robot selects the most promising candidate location $p^{*} \in C$, according to an exploration strategy,
\item[(d)] the robot reaches $p^{*}$, updates $p_{R}$, and restarts from (a).
\end{enumerate}
The above steps are repeated until no frontier is left and the map $M$ represents all the free space of $E$ reachable from the initial location of the robot.

The robot maintains a \emph{grid map} $M$ using a SLAM algorithm. We use GMapping~\cite{gmapping2007tro} in our experiments. Each cell of $M$ can be known or unknown and, in the former case, free or occupied. 
Given $M$, we identify the chains of free cells that are adjacent to at least an unknown cell. Each of such chains is a \emph{frontier} and the middle cell of each frontier is a candidate location. More precisely, a \emph{candidate location} is the cell that divides a frontier into two equal segments. Hence, given $M$, we identify a set $C$ of candidate locations. Each candidate location $p \in C$ is evaluated in step (c) according to a utility function $u(p)$ that combines distance and information gain (e.g., as in \cite{doi:10.1177/0278364902021010834,Basilico2011}). In particular ($\alpha \in [0,1]$ is a parameter that weights the two components),
\begin{equation}
\label{E:utility}
	u(p) = \alpha \cdot d(p) + (1-\alpha) \cdot i(p).
\end{equation}
In equation~(\ref{E:utility}), $d(p)$ is the \emph{distance} utility value:
	\begin{equation}
		\label{eqn:evaluation_utility_distance}
		d(p) = \frac{D_{\max} - D(p, p_{R})}{D_{\max}},
	\end{equation}
	\noindent where $D(p,p_{R})$ is the Euclidean distance between the current location of the robot $p_{R}$ and the candidate location $p$ and $D_{\max}$ is the maximum $D(p,p_{R})$ over all the candidate locations $p \in C$. In equation~(\ref{E:utility}), $i(p)$ is the \emph{information gain} utility value calculated as:
	\begin{equation}
		\label{eqn:evaluation_utility_ig}
		i(p) = \frac{I(p)}{I_{\max}},
	\end{equation}
	where $I(p)$ is the estimate of the amount of new (unexplored) area visible from $p$ (calculated as described in Section~\ref{S:SOL}) and $I_{\max}$ is the maximum value of $I(p)$ over all the candidate locations $p \in C$.	 
The next best candidate location $p^*$ is selected as follows:
\begin{equation}
p^* = \argmax_{p \in C} u(p).
\label{eqn:utility_func_max_problem}
\end{equation} 
Given a value of $\alpha$, $p^*$ represents the best balance between closeness and expected new area visible and, as such, is considered the best greedy choice for efficient exploration of the environment~\cite{4543637}. 

As we consider indoor environments, each candidate location $p \in C$ lies on a frontier and is in some partially observed room (defined below). The main idea of our method is to infer the possible geometrical shape (layout) of partially observed rooms, on the basis of structural features of the observed part of the environment. In particular, given the current grid map $M$, we retrieve its layout $\mathcal{L}$ identifying the rooms and labeling them as fully or partially observed. While the layout of fully observed rooms is known, that of partially observed rooms is predicted (see Section~\ref{sec:LAY}).
Then, the predicted layout of a partially observed room containing $p$ is used to provide an informed value for $I(p)$ in equation~(\ref{eqn:evaluation_utility_ig}) (see Section~\ref{S:SOL}).

\subsection{Retrieving the layout from grid maps}\label{sec:LAY}

\begin{figure*}[t!]
 \centering
    	\subfloat[Partial grid map.\label{fig:1A}]{ \includegraphics[width=0.2\textwidth]{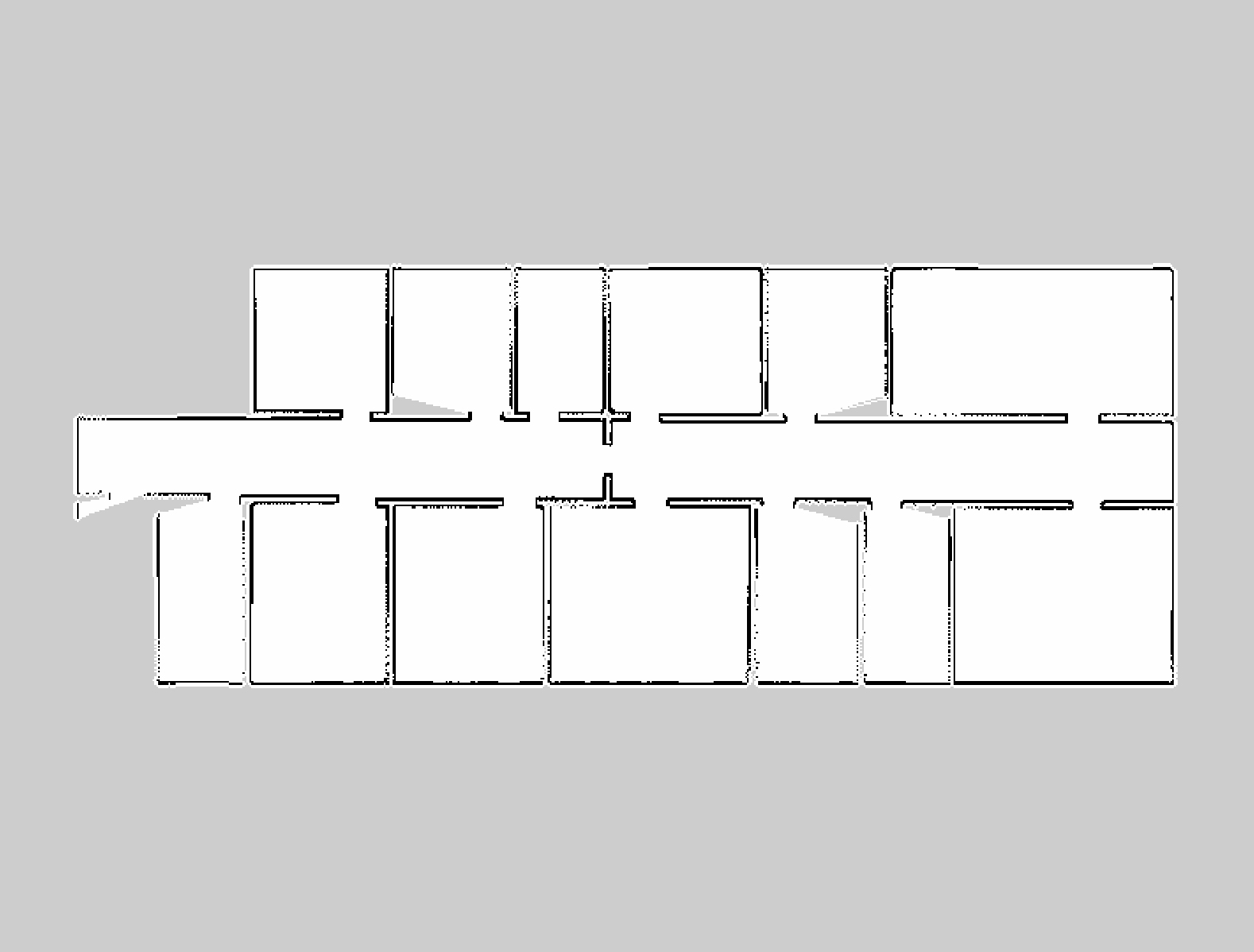}} 
	 \subfloat[Representative lines and faces.\label{fig:1B}]{ \includegraphics[width=0.2\textwidth]{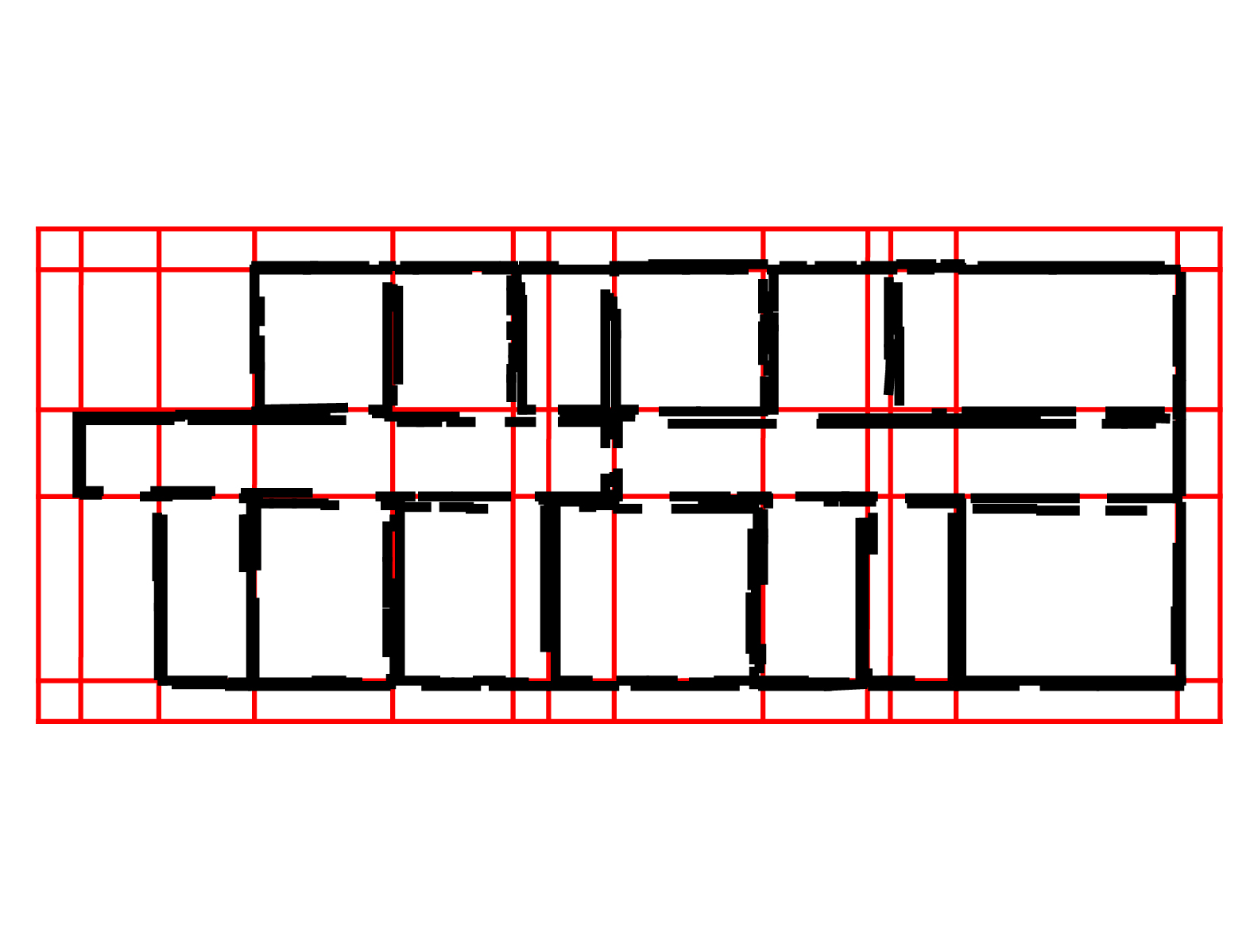}} 
    	\subfloat[Layout of fully observed rooms.\label{fig:1C}]{ \includegraphics[width=0.2\textwidth]{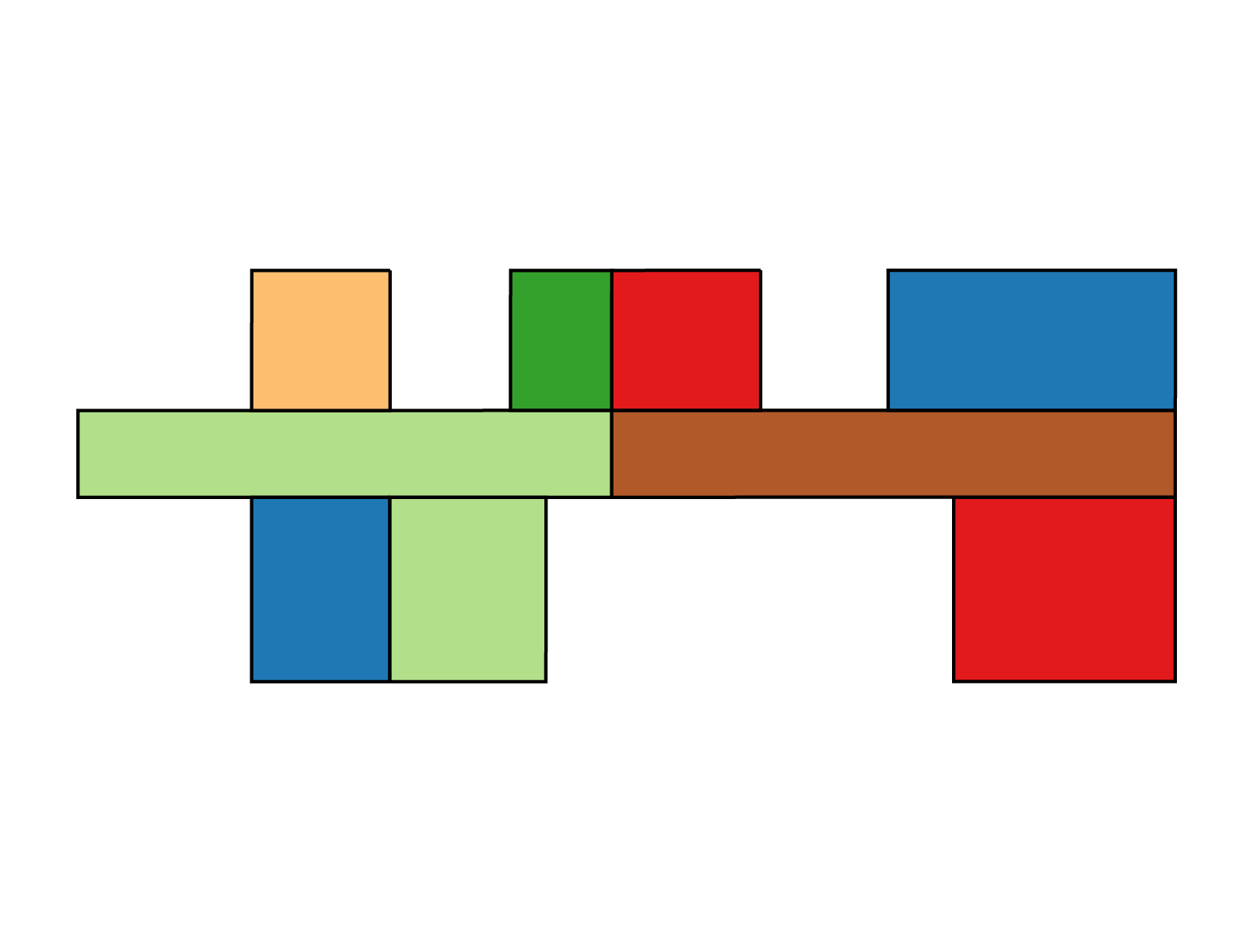}} 
    	\subfloat[Retrieved layout $\mathcal{L}$.\label{fig:1D}]{ \includegraphics[width=0.2\textwidth]{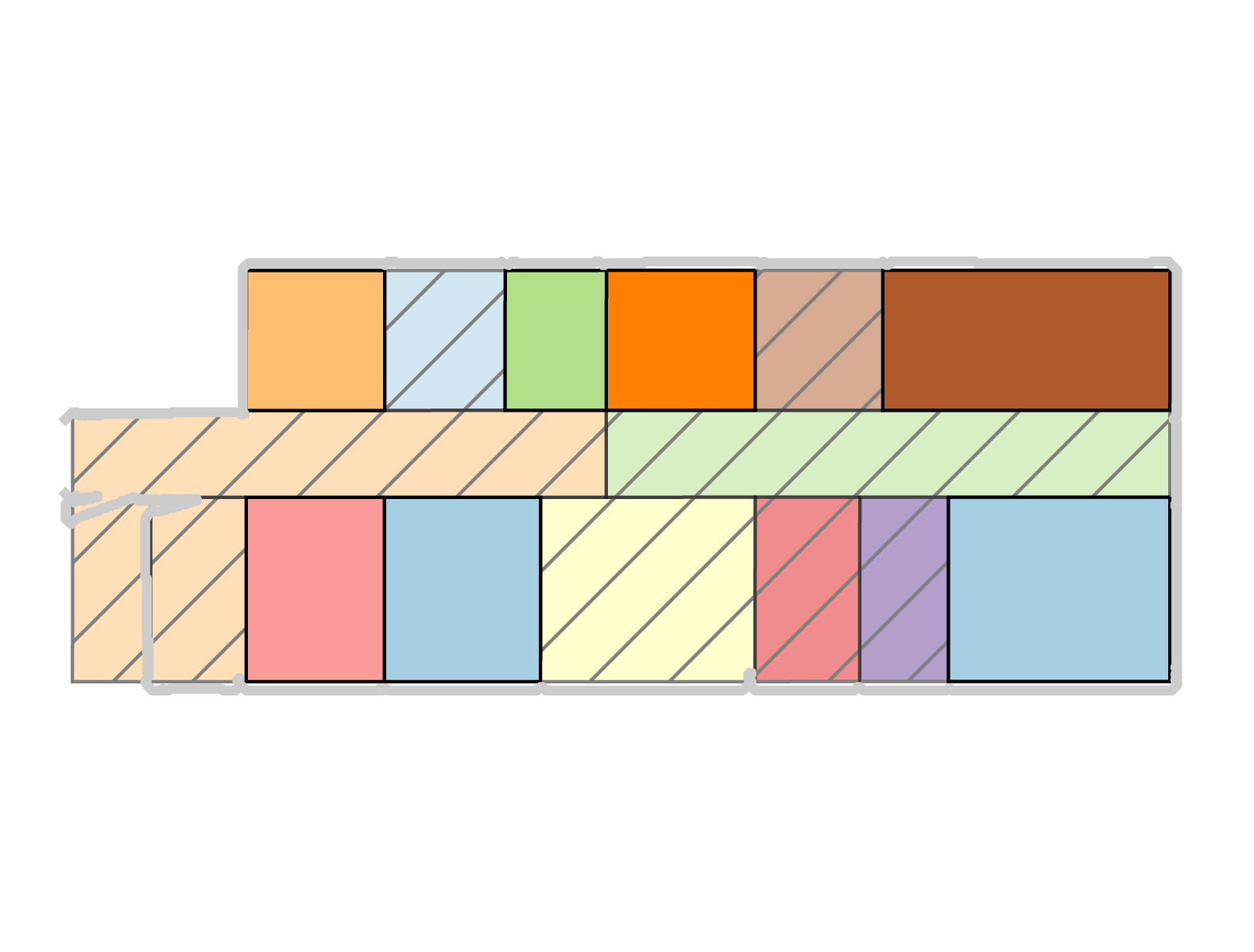}}   
    	\\	
 \caption{An example run of our approach for retrieving the layout starting from a partial grid map. In the map of Fig.\ref{fig:1A} $95\%$ of the area has been explored. In Fig. \ref{fig:1D}, the layout of partially observed rooms is dashed and the layout of fully observed rooms is solid. The map known to the robot is superimposed with gray lines. The same color code is used for layouts $\mathcal{L}$ in the rest of the paper. The fully explored map can be seen in \cite{IAS15}. \label{fig:EXE}} 
\end{figure*}

To retrieve the layout $\mathcal{L}$ of the environment starting from its partial grid map $M$ we use the method presented in~\cite{IAS15,ICRA19}. We provide here a summary of the algorithms using a running example (Fig. \ref{fig:EXE}). Please refer to the original papers for full details.

Our method starts from a partial grid map $M$ of the environment, like that of Fig. \ref{fig:1A}. From $M$, a set of edges is extracted and used to identify walls. 
Each wall is then associated to a \emph{representative line}, which indicates the direction of collinear (along the same direction), but possibly spatially separated, walls. Representative lines are reported in red in Fig. \ref{fig:1B}
and segment the environment into a set of \emph{faces}. 
Faces can be of three types: \emph{fully observed}, if their area has been completely observed in $M$, \emph{partially observed}, if their area has been partially observed in $M$, and \emph{unknown}, if no point of their area has been observed in $M$.

Then, the faces are clustered in groups, each one representing a room. We distinguish between \emph{fully observed} rooms, only composed of fully observed faces, and \emph{partially observed} rooms, also composed  of partially observed or unknown faces.
We start from identifying fully observed rooms, by clustering together fully observed faces that are adjacent, with their common edges not being walls in $M$. The polygon representing the layout of a room is obtained by merging the faces composing the room. An example of fully observed rooms identified from the partial grid map of Fig.~\ref{fig:1A} is in Fig.~\ref{fig:1C}.  

Then, we identify partially observed rooms by using information from representative lines, faces, and fully observed rooms. The idea is to find out the best set of faces (not belonging to any room) to form a room whose shape is ``consistent'' with the structure of the rest of the environment. For example, if one side of a room is bounded by a corridor, the opposite side of that room likely shares a wall with adjacent rooms along the same corridor.
Practically, we start from a partially observed face $f$ that contains a frontier and we iteratively consider all the sets of adjacent faces (fully observed, partially observed, and unknown, which have been not clustered in any room) that can form a room (e.g., they must be connected). 
For each set of faces $F$, we calculate an objective function $\Phi(F \cup \{ f\})$ composed of three terms that evaluate the consistency of the predicted room shape wrt that of fully observed rooms, the simplicity of the predicted room shape, and the number of walls of the predicted room shape, respectively. 
Finally, the set of faces $F^{*} \cup \{f\}$ that maximizes $\Phi(F \cup \{ f\})$ is associated with the partially observed room. The polygon representing the predicted layout of the room is found by merging the faces $F^{*} \cup \{ f\}$. 
An example of the predicted layout of partially observed rooms is in Fig \ref{fig:1D}. The retrieved layout $\mathcal{L}=\{ r_{1}, r_{2}, \ldots\}$ is eventually composed of the layout of both fully observed and partially observed rooms.

A particular situation is encountered when a partially observed room is at the border of the map $M$. In this case, one or more sides of the room are not bounded by any representative line derived from $M$ and the layout of the room cannot be predicted, as we cannot exploit any knowledge for making such estimation.
When this happens, we label the room as containing an \emph{open frontier} and we highlight the corresponding edges in red, as in the first three examples of Fig.~\ref{fig:INC1} (discussed later). This particular situation usually occurs at early stages of exploration, where only a limited portion of the environment has been explored.

\subsection{Expected information gain I(p)}\label{S:SOL}

\begin{figure}[b!]
	\centering
	\subfloat[][Candidate location $p$ (blue dot).\label{fig:estimate_inf_gain_wt_1}]{	\includegraphics[ width=0.22\linewidth]{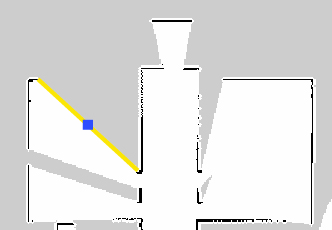}} \hspace{0.2cm}
	\subfloat[][$I(p)$ without using $\mathcal{L}$.\label{fig:estimate_inf_gain_wt_2}]{ \includegraphics[width=0.22\linewidth]{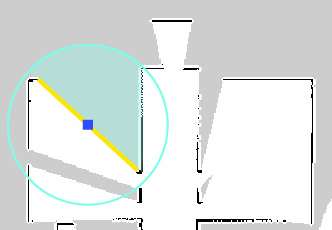}} 
	\subfloat[][Layout $\mathcal{L}$. \label{fig:estimate_inf_gain_wt_3}]{ \includegraphics[width=0.22\linewidth]{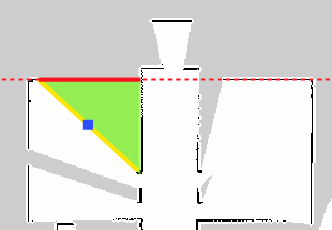}} \hspace{0.2cm}
	\subfloat[][$I(p)$ using $\mathcal{L}$.\label{fig:estimate_inf_gain_wt_4}]{ \includegraphics[width=0.22\linewidth]{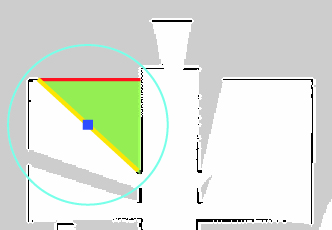}}
	\caption{An example of how $I(p)$ is calculated without (Fig. \ref{fig:estimate_inf_gain_wt_2}) and with (Fig. \ref{fig:estimate_inf_gain_wt_4}) the knowledge of $\mathcal{L}$. Fig. \ref{fig:estimate_inf_gain_wt_3} shows the representative line (red, dashed) of the wall used to predict the layout of the partially observed room in $\mathcal{L}$.}
	\label{fig:estimate_inf_gain_with_floorplan}
\end{figure}

In this paper, we originally exploit the retrieved layout $\mathcal{L}$ to calculate $I(p)$ in equation (\ref{eqn:evaluation_utility_ig}), namely to estimate the amount of unexplored area visible from a candidate location $p$ (see Fig.~\ref{fig:estimate_inf_gain_wt_1}).
The mainstream approaches for calculating $I(p)$ measure the maximum visible area from $p$ given the footprint of the robot's laser range scanner (as done, e.g., in~\cite{doi:10.1177/0278364902021010834,Basilico2011}) or the length of the frontier on which $p$ lies (as partially done, e.g., in~\cite{Tovar2006}). These approaches are designed for settings where no knowledge about the unobserved part of the environment is taken into account. Fig.~\ref{fig:estimate_inf_gain_wt_2} shows an example in which $I(p)$ is calculated as the area of the maximum number of unknown cells that can be perceived by the laser range scanner from $p$. This estimate is optimistic and implicitly assumes that the area beyond the frontier on which $p$ is located is free.

In our approach, we calculate $I(p)$ as follows. Given a map $M$ and its retrieved layout $\mathcal{L}$, the walls identified in $\mathcal{L}$ (corresponding to the edges of the polygons representing the rooms) are projected on $M$  as obstacles, thus obtaining a new map $M^{\mathcal{L}}$. Given a cell $p \in M$, we find the corresponding cell $p^{\mathcal{L}} \in M^{\mathcal{L}}$. Then, for each unknown cell $c \in M$ that is within the footprint of the laser range scanner when the robot is in $p$, we find the corresponding $c^{\mathcal{L}} \in M^{\mathcal{L}}$. The cell $c$ contributes to calculating the expected area $I(p)$ visible from $p$ when both the following conditions are satisfied:
\begin{itemize}
	\item $c^{\mathcal{L}}$ is free,
	\item $c^{\mathcal{L}}$ is visible from $p^{\mathcal{L}}$ in $M^{\mathcal{L}}$, namely the line segment connecting their centers does not touch any obstacle cell in $M^{\mathcal{L}}$.
\end{itemize}
Eventually, given the cells $c$ that satisfy the above conditions, $I(p)$ is calculated by summing the areas of those cells. 
Fig.~\ref{fig:estimate_inf_gain_wt_4} shows an example in which the method just described is used to calculate $I(p)$. It is interesting to contrast it with Fig.~\ref{fig:estimate_inf_gain_wt_2}. Although it is a sort of informed variant of the classical frontier-based exploration approaches, the proposed method provides benefits to the performance of exploration also when the retrieved layout $\mathcal{L}$ and the grid map $M^{\mathcal{L}}$ in which it is embedded are inaccurate, as we show in the next section. 
In case of open frontiers, where the information about $\mathcal{L}$ cannot be used to estimate $I(p)$, the maximum area that can be perceived from $p$ is considered (as in~\cite{doi:10.1177/0278364902021010834,Basilico2011}).

\subsection{Early stopping of exploration}\label{s:es}

Exploration missions are usually performed until the entire area of the environment is mapped by the robot. As a consequence, classical exploration techniques as \cite{doi:10.1177/0278364902021010834,Basilico2011} often result in the following behavior: at the beginning of exploration, the robot quickly increments its map $M$ by visiting locations with high information gain. However, it usually leaves behind small scattered frontiers across different rooms, as in the example of Fig.~\ref{fig:EXE}. 
Hence, in the final stages of the exploration process (e.g., at $90$ or $95\%$ of the total area explored), the robot has to reach all such remaining frontiers and perceive the environment from there. These residual frontiers are particularly costly to visit, being usually in rooms far away one from each other, and often result in small information gains, as they usually represent small gaps like corners.

The retrieved layout $\mathcal{L}$ can be used to estimate the missing parts of partially observed rooms and to automatically fill the small gaps without actually observing them. Considering the example of Fig. \ref{fig:EXE}, our method is able to provide a correct estimate of the shape of the area visible from all of the remaining frontiers. In order to exploit this feature, we introduce a criterion for stopping the exploration early, which is based on $\mathcal{L}$. More precisely, \emph{Early Stopping} (ES) ends the exploration if the estimated unexplored area visible from all the current candidate locations in $C$ is less than a threshold. When ES is triggered, we consider the exploration complete and discard the remaining frontiers, as we can easily predict the area that can be perceived from them.

%% file: 04-Exp.tex
\section{Experimental Evaluation}\label{sec:EXP}

In this section, we evaluate the proposed approach in exploring large-scale simulated buildings. 

\subsection{Experimental setting}
We implemented our method in ROS, using the ROS navigation stack. Explorations are performed in $10$ large-scale buildings (from \SI{1000}{\square\meter} to \SI{3500}{\square\meter}) simulated in Stage\footnote{\url{http://wiki.ros.org/stage}}, using a simulated robot equipped with a laser range scanner with a field of view of $180^{\circ}$ and a range of \SI{6}{\meter}. We perform, for each environment, $10$ runs using our method for estimating the information gain $I(p)$ (``with $\mathcal{L}$'') and using a baseline method similar to those of \cite{doi:10.1177/0278364902021010834,Basilico2011}, which is representative of those currently most used in the literature (``no $\mathcal{L}$''), in which the information gain $I(p)$ is calculated as in Fig.~\ref{fig:estimate_inf_gain_wt_2}. We measure, as exploration progresses, the percentage of \emph{explored area} ({\it exp}), namely the percentage of free area of $E$ mapped in $M$, as a function of the \emph{time}. We compute mean and standard deviation (over all the runs) of the time required to perform a full exploration for each environment (namely time at {\it exp}~$=100\%$). The starting locations of the robot are at the entrances of the buildings and are fixed for all the runs.  
The maps obtained in different runs are slightly different from each other, because of the noise introduced in the simulation (translational error up to $\SI[per-mode = symbol]{0.01}{\meter\per\meter}$ and rotational error up to $\SI[per-mode = symbol]{2}{\degree\per\radian}$), resulting in different frontiers being detected and, ultimately, in different choices being made by the robot.

The layout $\mathcal{L}$ is retrieved online starting from the grid map $M$ provided by the ROS implementation\footnote{\url{http://wiki.ros.org/gmapping}} of GMapping~\cite{gmapping2007tro}. 
At the beginning of the exploration, as only a few rooms have been fully observed, the predicted layout of partially observed rooms could be highly inaccurate, and several open frontiers could be detected. Note that, differently from methods like 
\cite{LearnedMap}, we do not require any prior data to learn a model. With the progression of the exploration, however, $\mathcal{L}$ becomes more stable and accurate. 
The computation of $\mathcal{L}$ takes few seconds at early stages of exploration and less than \SI{10}{\second} at final stages, and is performed by a dedicated ROS node so that an updated version of $\mathcal{L}$ is always available without introducing delays. After several preliminary tests, we set $\alpha = 0.5$ in (\ref{E:utility}) to equally balance distance and information gain and guarantee a good overall performance in exploration.

Exploration runs are concluded when no frontier is left and the set of candidate locations $C$ is empty. However, when we employ the ES variant of Section \ref{s:es}, the exploration is stopped when the estimated unexplored area visible from all the current candidate locations in $C$ is less than \SI{1}{\square\meter}. We set this threshold to a rather conservative value. If increased, the exploration process could terminate earlier, at the cost of possible inaccuracies in estimating the unobserved parts of the environment. This issue is discussed later.

\subsection{Experimental results}

\begin{table}[t!]
\centering
\resizebox{0.6\linewidth}{!}{%
	\begin{tabular}{cccccccc}
	\toprule
	 \multicolumn{2}{c}{no $\mathcal{L}$} & \multicolumn{2}{c}{with $\mathcal{L}$} & \multicolumn{2}{c}{with $\mathcal{L}$ + ES} & \multicolumn{2}{c}{gain} \\
 $t$  & $\sigma$   &  $t$      & $\sigma$   & $t$  & $\sigma$   & with $\mathcal{L}$ & with $\mathcal{L}$ + ES      \\
        	 \midrule
	 3440 & 313 & 3090 & 316 & 3003 & 321 & 10.1\% & 12.8\% \\  			\bottomrule
	\end{tabular}
	}
\caption{Exploration results over $10$ runs in $10$ simulated large-scale buildings. $t$ (in \SI{}{\second}) is the average time and $\sigma$ is the corresponding standard deviation. ``no $\mathcal{L}$'' is the baseline method, ``with $\mathcal{L}$'' is our proposed method, and ``with $\mathcal{L}$ + ES'' is the variant of our method with ES. The percentage gain of our method is reported in the last two columns. \label{tab:results}}
\end{table}

\begin{figure}
\centering
\includegraphics[width=0.85\textwidth]{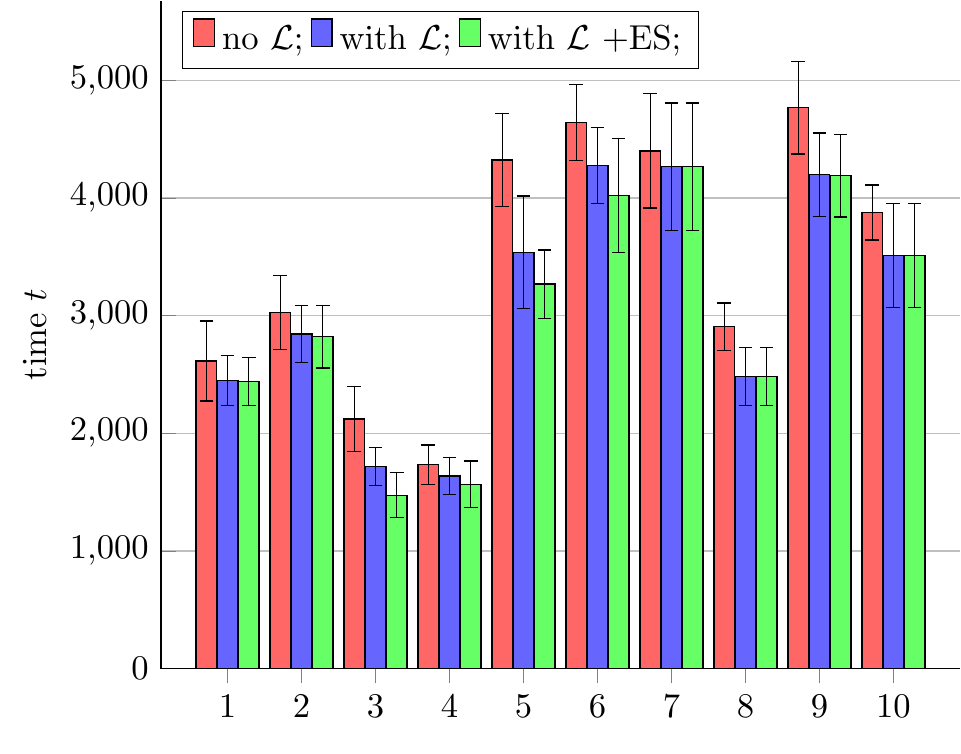}
 \caption{Average exploration time (and standard deviation) for all the $10$ environments (on the $x$ axis) with {\it exp} $=100\%$.\label{fig:BAR}} 
\end{figure}

Table \ref{tab:results} reports the average time required for performing $10$ complete exploration runs in all the $10$ large-scale environments. The time includes retrieving the layout $\mathcal{L}$ at each stage. The use of $\mathcal{L}$ for estimating the information gain brings a speedup of the total exploration time of approximatively the $10\%$. More detailed results on the average time (over $10$ runs) required to completely explore each one of the $10$ environments are reported in Fig. \ref{fig:BAR}. 
For all $10$ environments, our method results in a speedup of exploration. The use of ES further decreases the time for almost all of the environments.

 \begin{figure*}[t!]
 \centering
 \captionsetup[subfigure]{labelformat=empty}
    	\subfloat{ \includegraphics[trim={0 2cm 0cm 2cm},clip,width=0.19\textwidth]{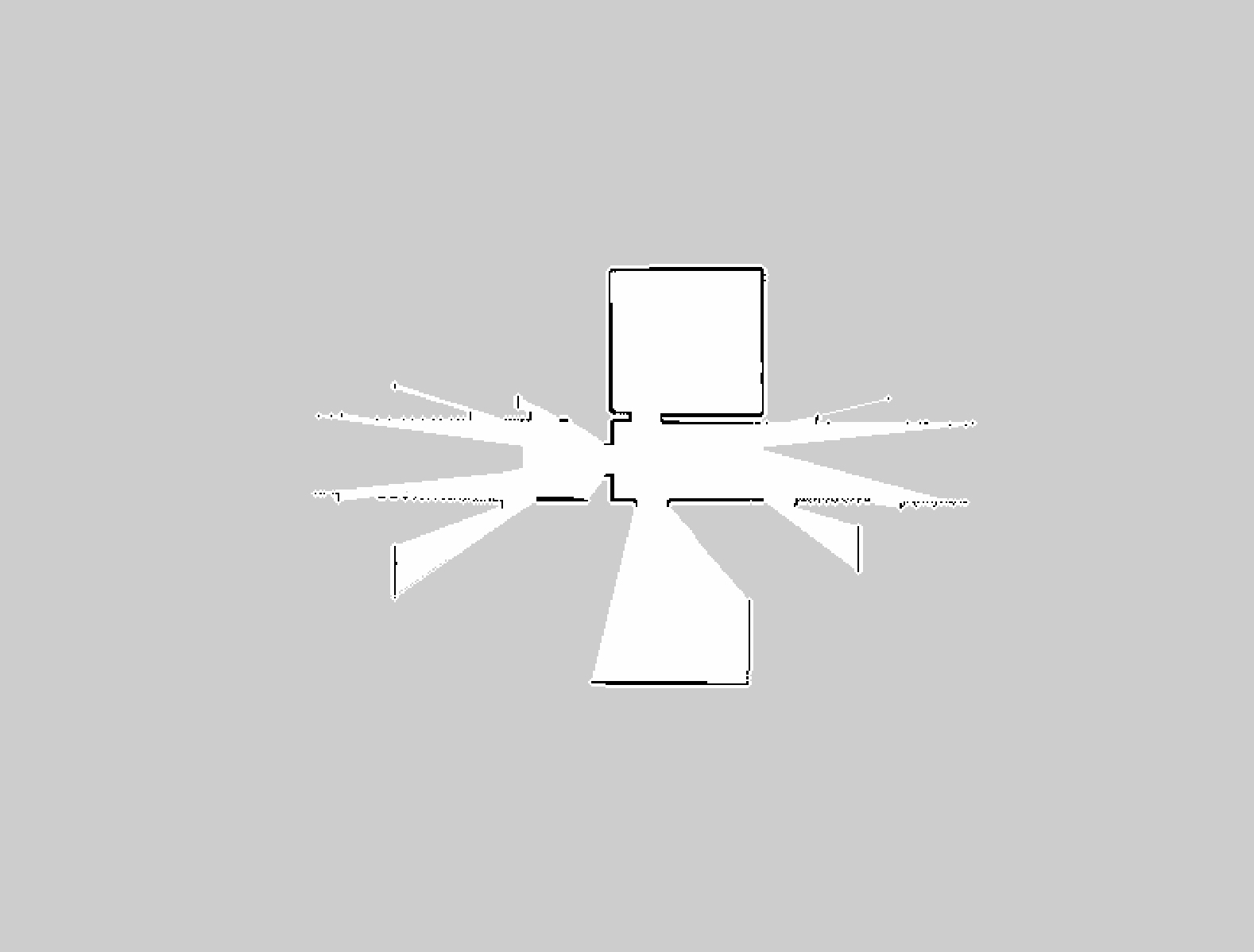}}
    	\subfloat{ \includegraphics[trim={0 2cm 0cm 2cm},clip,width=0.19\textwidth]{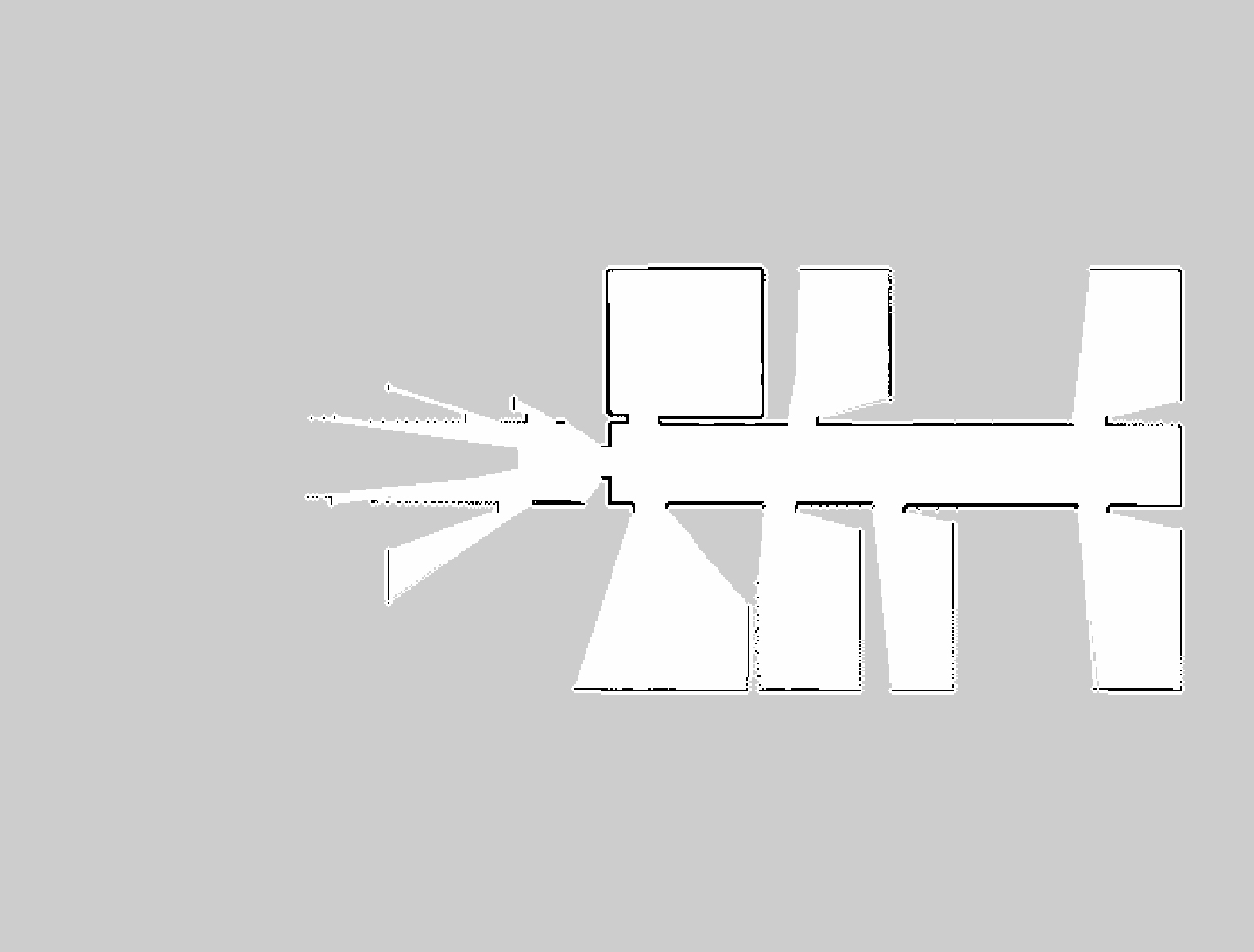}}   
		\subfloat{ \includegraphics[trim={0 2cm 0cm 2cm},clip,width=0.19\textwidth]{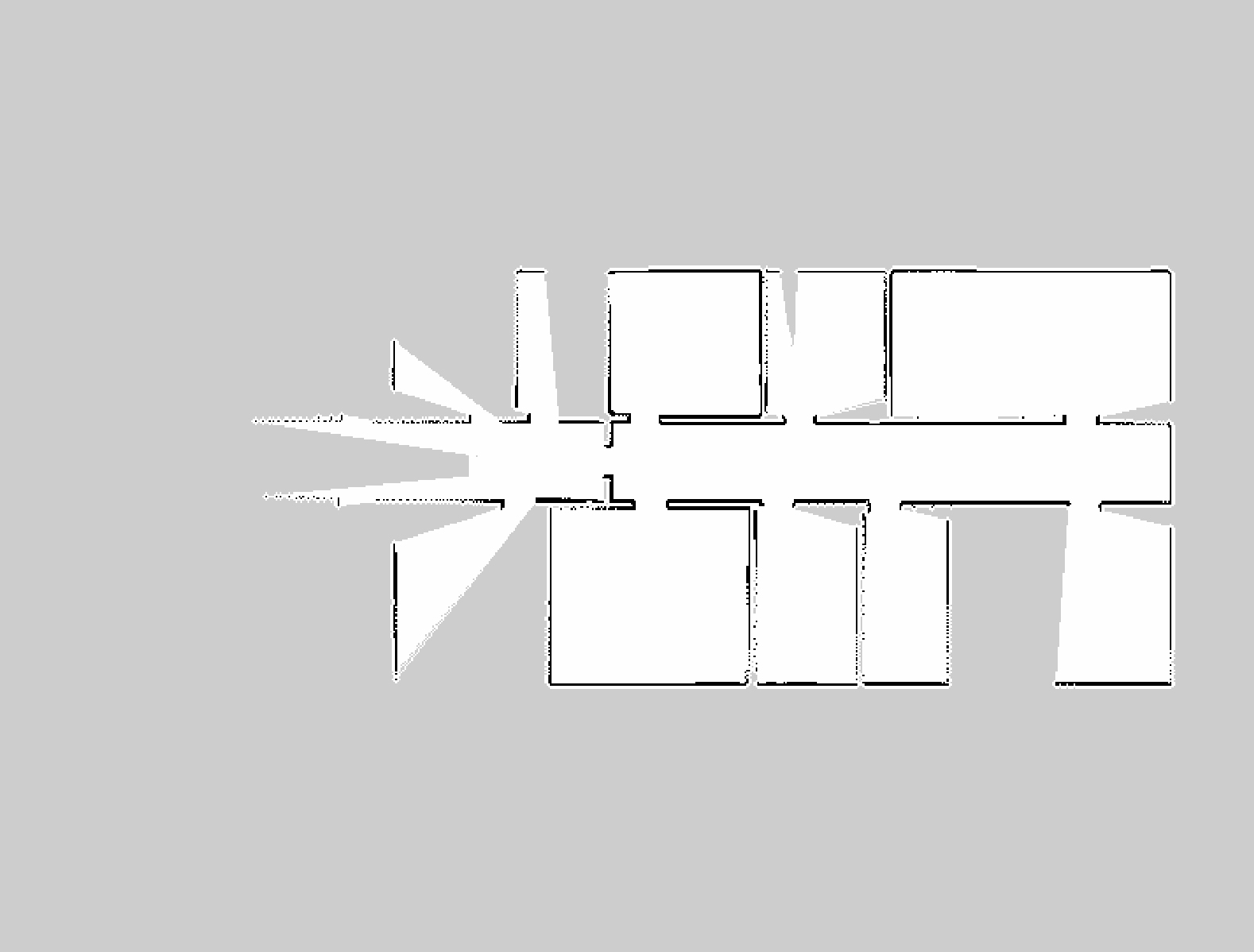}}
    	\subfloat{ \includegraphics[trim={0 2cm 0cm 2cm},clip,width=0.19\textwidth]{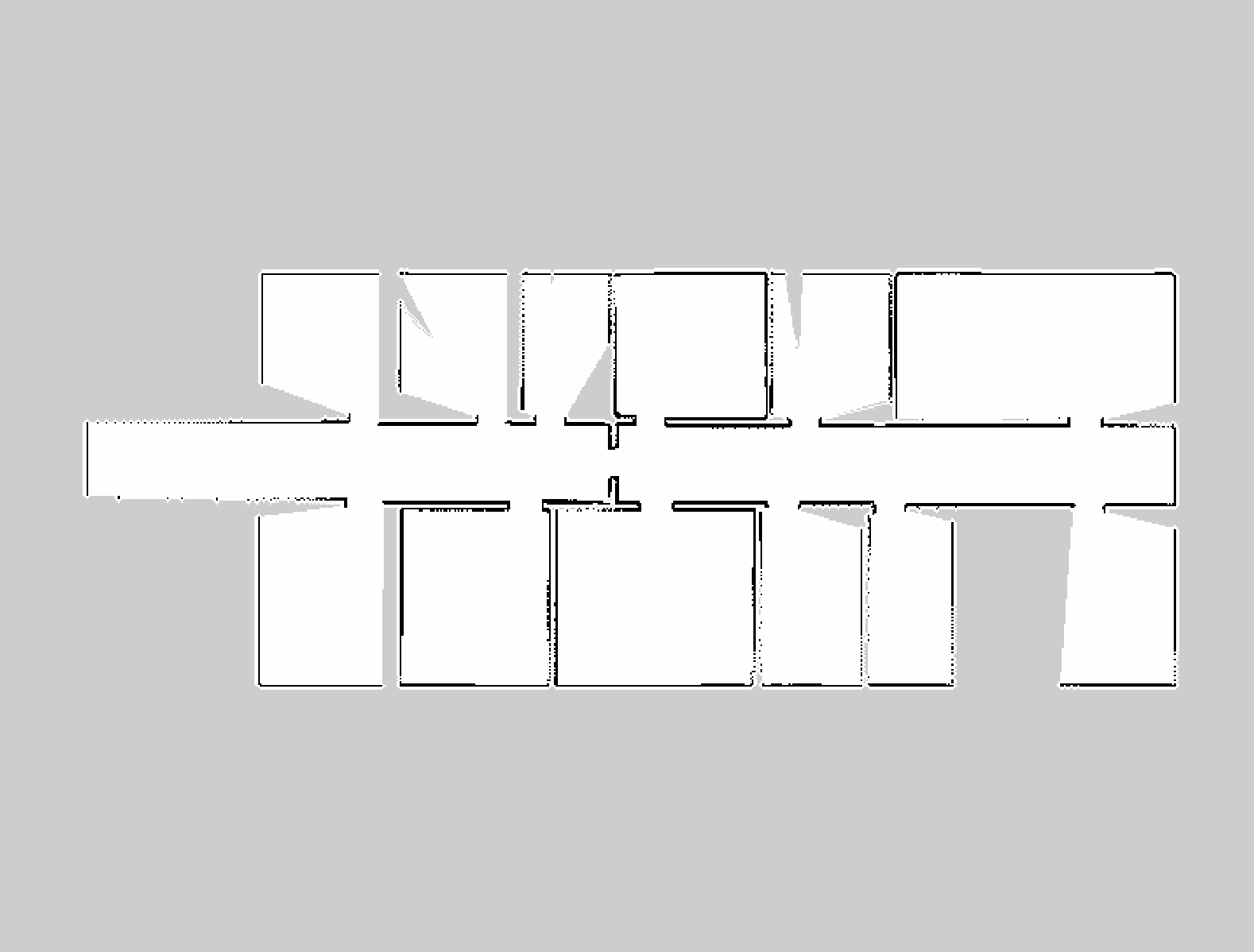}}
    	\subfloat{ \includegraphics[trim={0 2cm 0cm 2cm},clip,width=0.19\textwidth]{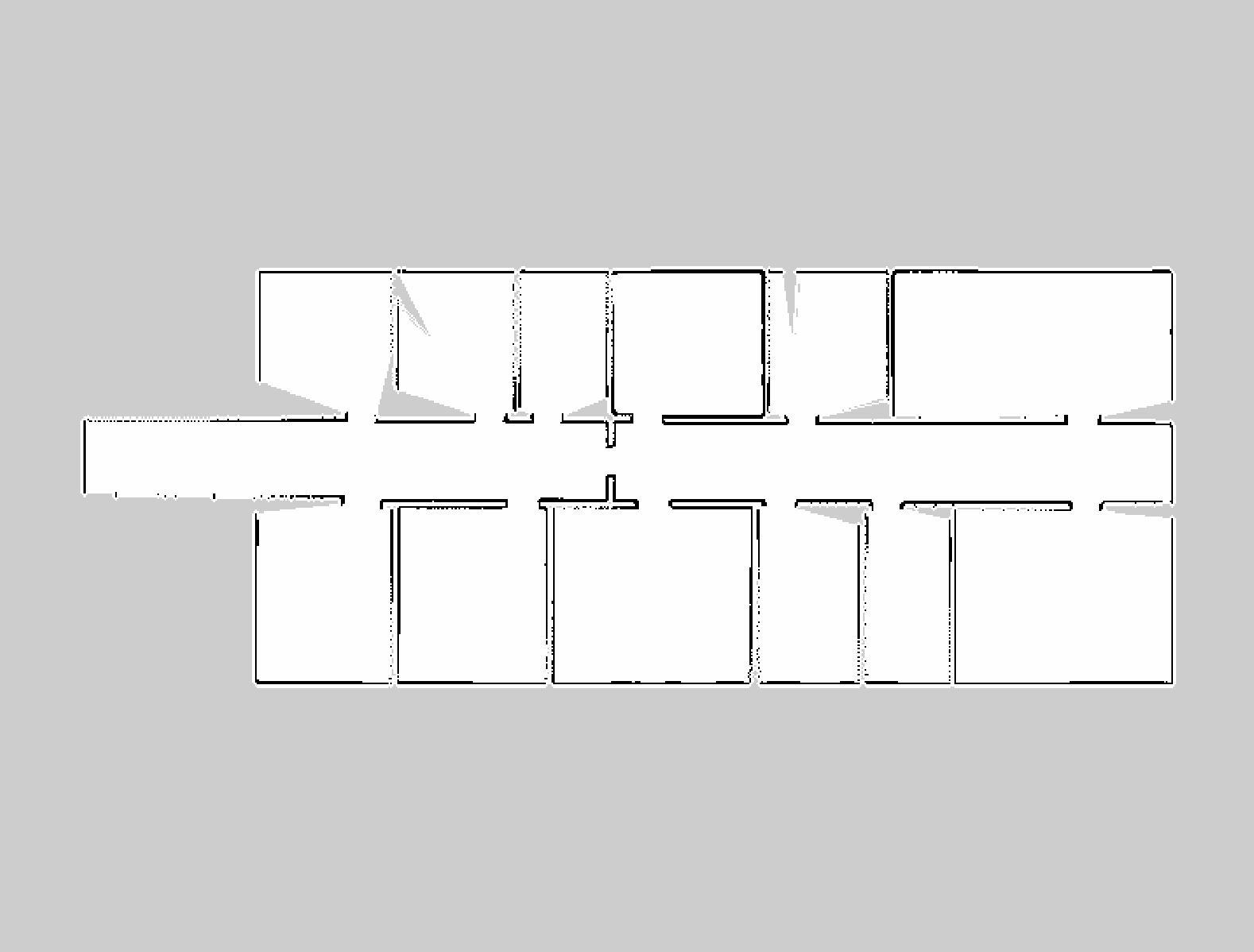}} \\
    	    	
    	\subfloat[{\it exp} $=20\%$ \label{fig:INC_1A}]{ \includegraphics[trim={0 2cm 0cm 2cm},clip,width=0.19\textwidth]{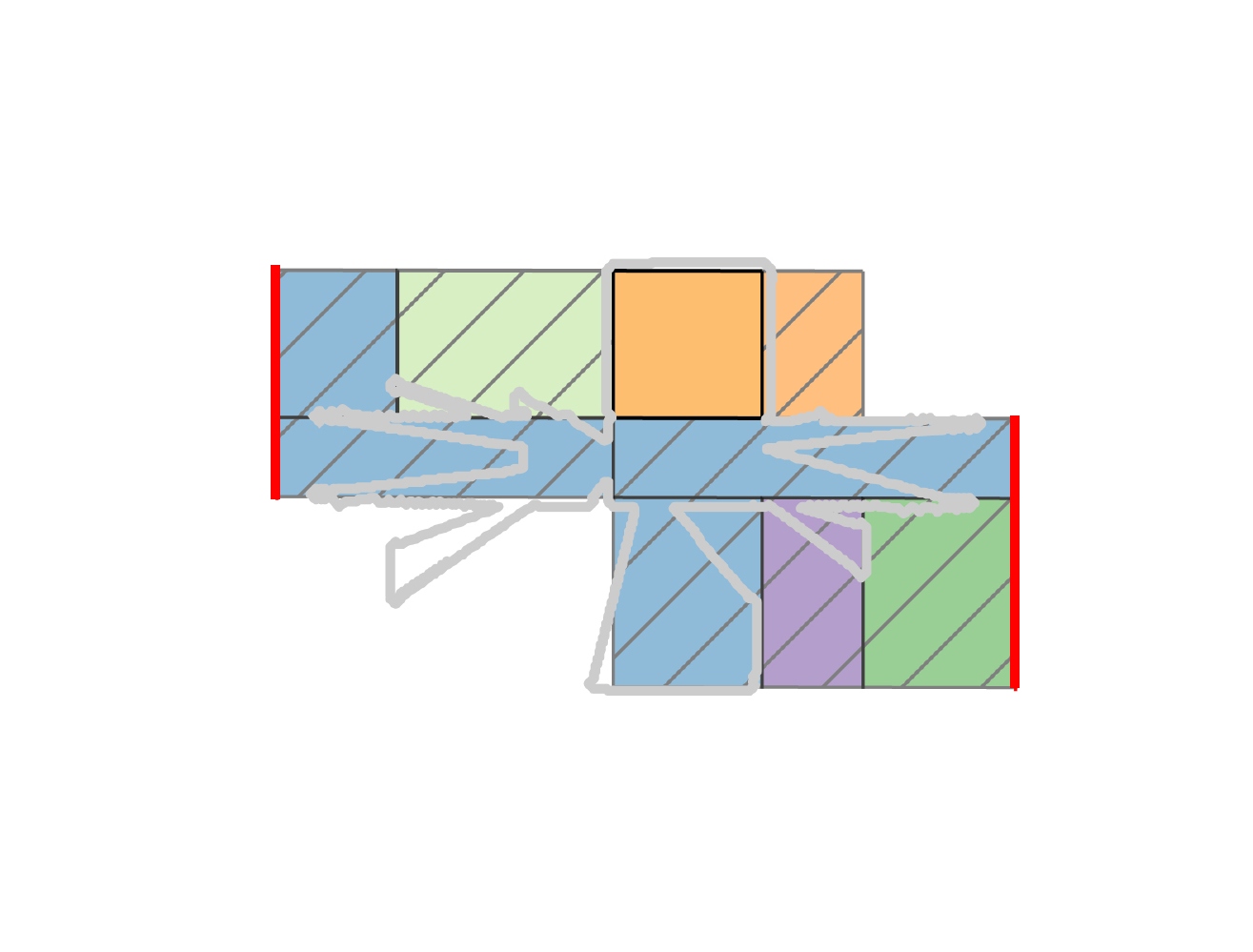}}
    	\subfloat[{\it exp} $=40\%$\label{fig:INC_1B}]{ \includegraphics[trim={0 2cm 0cm 2cm},clip,width=0.19\textwidth]{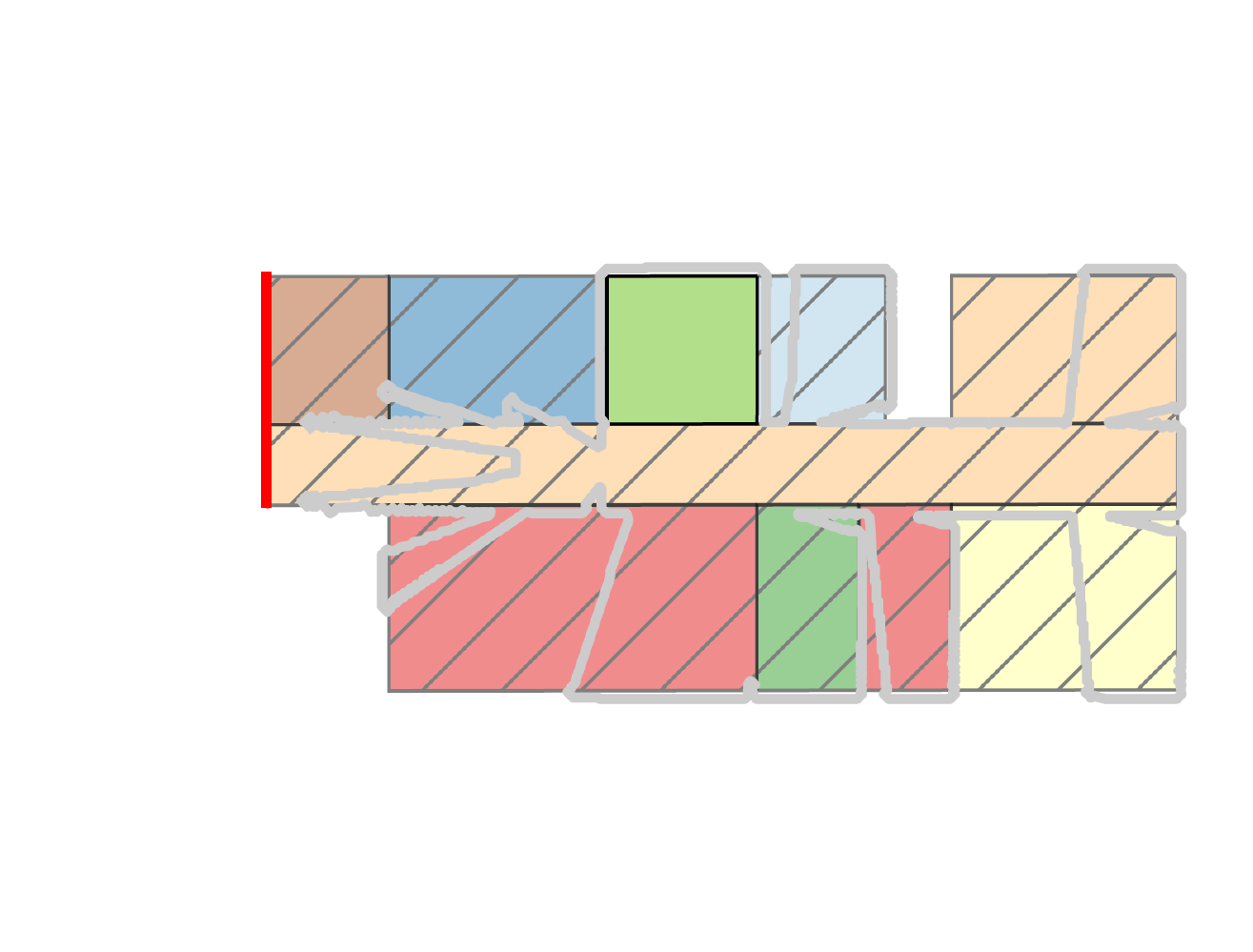}}   	
		\subfloat[{\it exp} $=60\%$ \label{fig:INC_1C}]{ \includegraphics[trim={0 2cm 0cm 2cm},clip,width=0.19\textwidth]{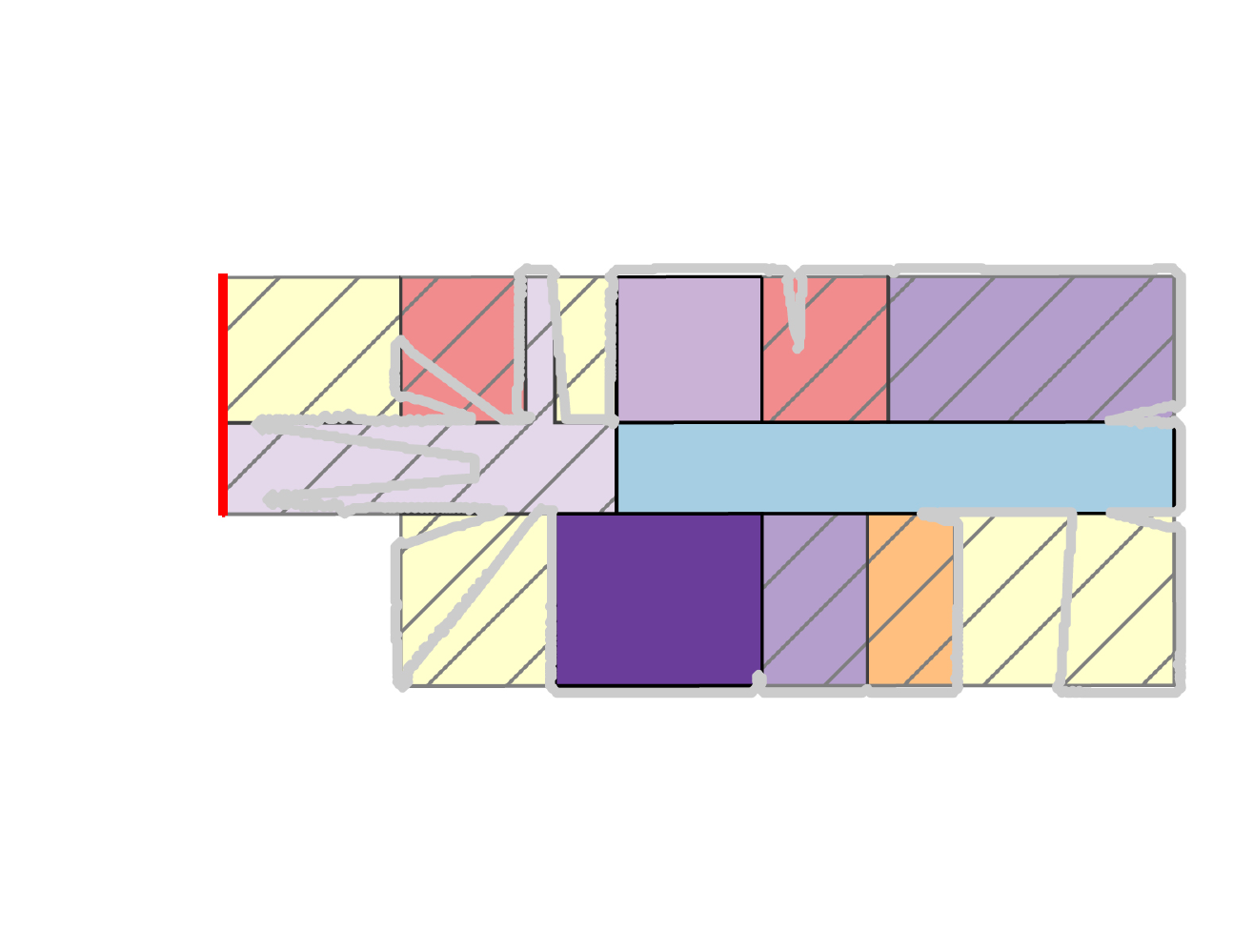}}
    	\subfloat[{\it exp} $=80\%$ \label{fig:INC_1D}]{ \includegraphics[trim={0 2cm 0cm 2cm},clip,width=0.19\textwidth]{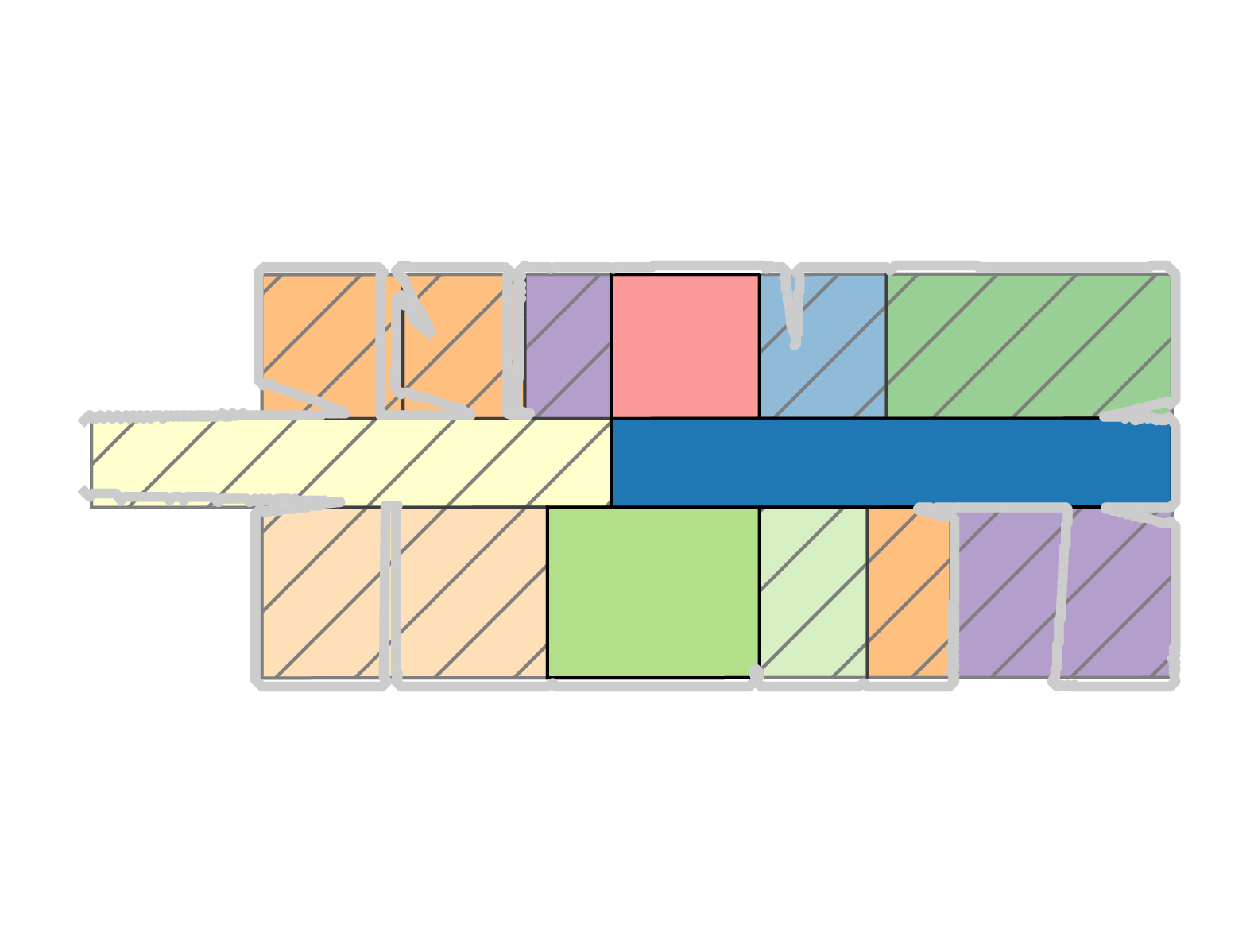}}
    	\subfloat[{\it exp} $=90\%$ \label{fig:INC_1E}]{ \includegraphics[trim={0 2cm 0cm 2cm},clip,width=0.19\textwidth]{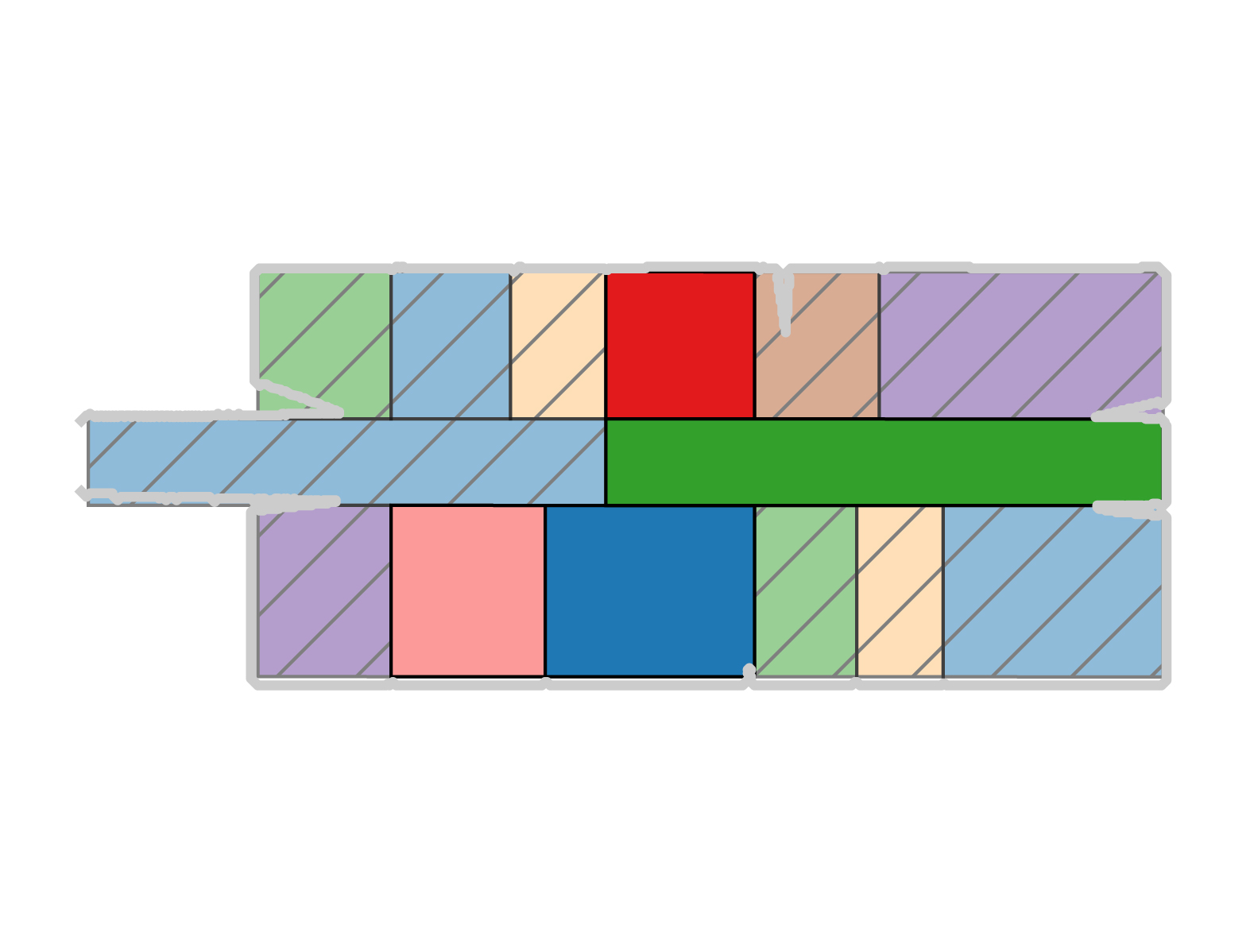}}	\\

\caption{Example of application of our approach to reconstruct the layout $\mathcal{L}$ from partial grid maps $M$ at different exploration stages of the same environment. The layout obtained in the same environment with {\it exp}~$=95\%$ is in Fig. \ref{fig:EXE}. \label{fig:INC1}} 
\end{figure*}

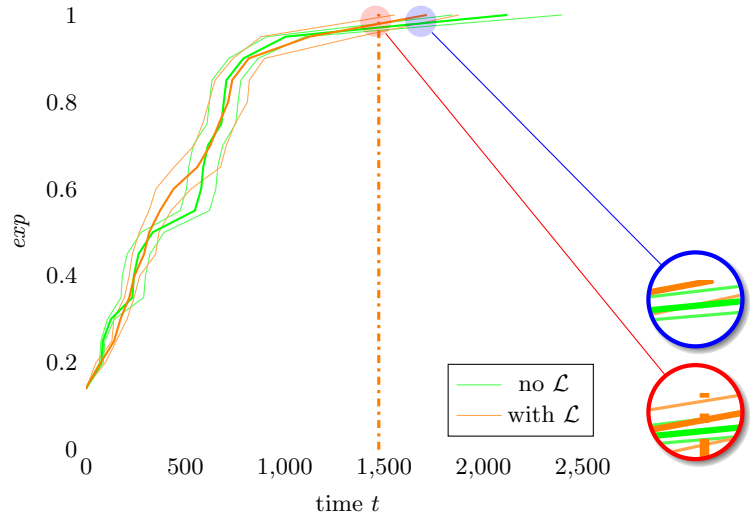
\begin{figure}[t]
\centering
\begin{tikzpicture}[spy using overlaysshadow={
           magnification=3, 
    size=1.5cm, 
    connect spies}
]
	\begin{axis}[
	width=0.7\linewidth, 
	ymajorgrids=true,
	grid=major,
	grid style=dashed,
	xlabel=time $t$, 
	ylabel={\it exp},
	xmin = 0,
	ymin = 0,
	ymax = 1,
	legend pos = south east,
	]

	\addplot[name path=prior_up,color=green!70]
	table[x expr=\thisrow{Time} - \thisrow{Std Deviation},y=ExploredArea,col sep=comma] {freiburg79_no_floorplan_ExploredAreaTime.csv};
	\addlegendentry{no $\mathcal{L}$}
	\addplot[name path=prior_up,color=orange!70]
	table[x expr=\thisrow{Time} - \thisrow{Std Deviation},y=ExploredArea,col sep=comma] {freiburg79_layout_reconstruction_ExploredAreaTime.csv};
	\addlegendentry{with $\mathcal{L}$}

	\addplot[name path=prior_down,color=green!70]
	table[x expr=\thisrow{Time} + \thisrow{Std Deviation},y=ExploredArea,col sep=comma] {freiburg79_no_floorplan_ExploredAreaTime.csv};
	\addplot[green!50,fill opacity=0.5] fill between[of=prior_up and prior_down];
	\addplot[name path=prior_down,color=orange!70]
	table[x expr=\thisrow{Time} + \thisrow{Std Deviation},y=ExploredArea,col sep=comma] {freiburg79_layout_reconstruction_ExploredAreaTime.csv};
	\addplot[orange!50,fill opacity=0.5] fill between[of=prior_up and prior_down];

\addplot [color = green,thick] table [x=Time, y=ExploredArea, col sep=comma] {freiburg79_no_floorplan_ExploredAreaTime.csv};

\addplot [color = orange,thick] table [x=Time, y=ExploredArea, col sep=comma] {freiburg79_layout_reconstruction_ExploredAreaTime.csv};
\addplot [ycomb,very thick,dashdotted,color = orange] table [x =Time,y expr=2,col sep = comma]{freiburg79_layout_reconstruction_closeMapPoint.csv};

\begin{scope}
\spy [red,size=1.25cm] on (3.85cm,5.7cm) in node at (8.1cm,0.5cm);
\spy [blue,size=1.25cm] on (4.45cm, 5.7cm) in node at (8.1cm,2cm);
\end{scope}
          \end{axis}
    \end{tikzpicture}
 \caption{The progression of the explored area (averaged over runs) as a function of time for the environment of Fig.~\ref{fig:INC1}. Early stopping is indicated as a dashed vertical line.\label{fig:PROGR}} 
\end{figure}

We look at the progression of the exploration during one run performed in the environment of Fig. \ref{fig:EXE}. Fig. \ref{fig:INC1} reports the grid maps and the corresponding retrieved layouts $\mathcal{L}$ at different exploration stages. $\mathcal{L}$ becomes more stable and accurate as the exploration progresses. Reliable estimates of $I(p)$ can be obtained at {\it exp}~$=60\%$ and even at {\it exp}~$=20\%$. Fig. \ref{fig:PROGR} reports the explored area (averaged over $10$ runs) as a function of time. Our method performs similarly to the baseline until the explored area is around $90\%$, when we are able to provide a better estimate of $I(p)$ and, consequently, to speed up the final part of the exploration. For this example, ``with $\mathcal{L}$'' has a gain of $19.1\%$ over ``no $\mathcal{L}$'', while ``with $\mathcal{L}$ + EP'' has a gain of $30.5\%$.

\begin{figure}
 \centering
    	\subfloat[Partial grid map.\label{fig:A}]{ \includegraphics[width=0.35\textwidth]{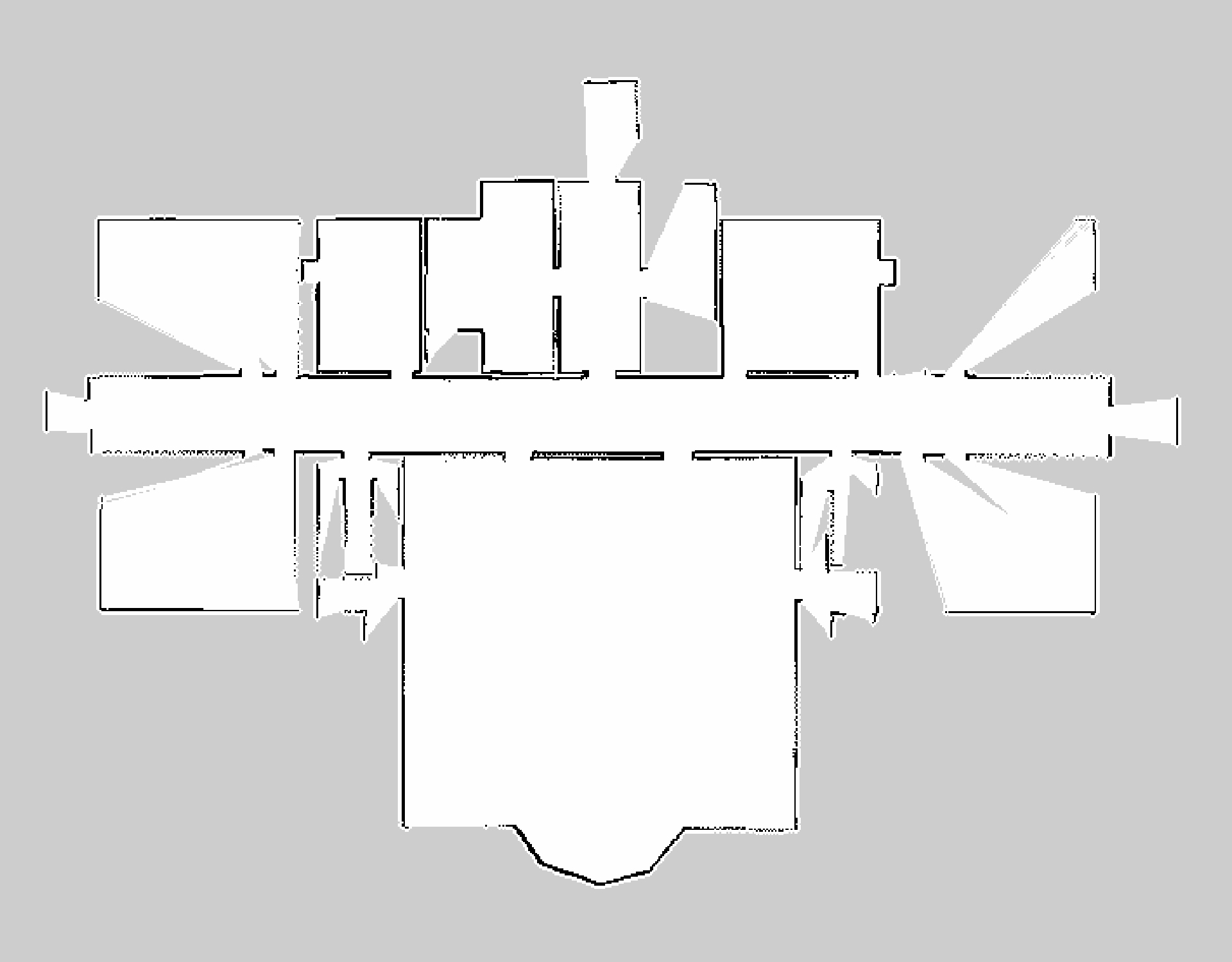}}
    	\subfloat[Reconstructed layout.\label{fig:B}]{ \includegraphics[width=0.35\textwidth]{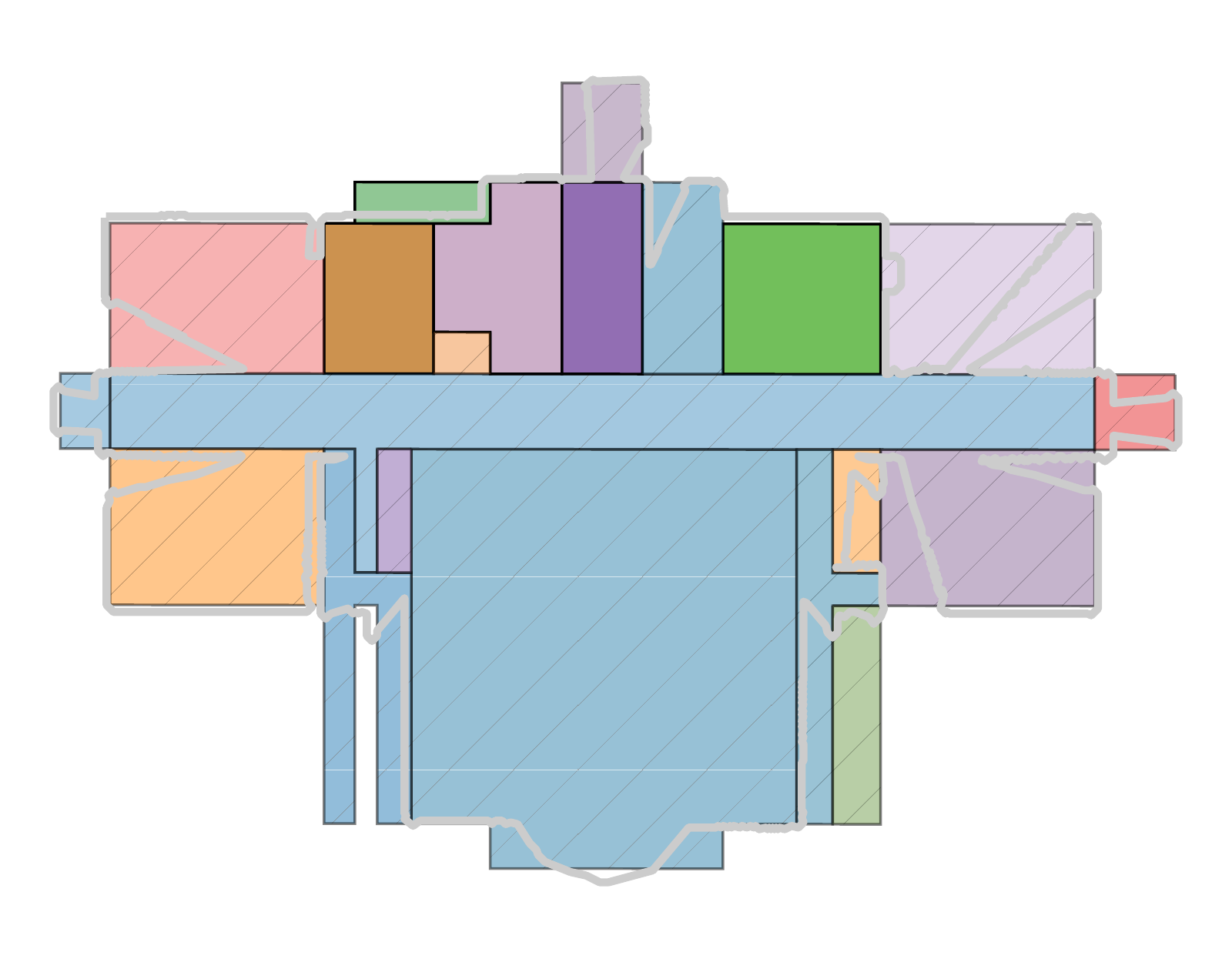}}   	
 \caption{An example of reconstructed layout $\mathcal{L}$ obtained at {\it exp} $=80\%$. Our method is able to predict the shape of the complete environment, so that exploration could be considered complete.\label{fig:EXEP1}} 
\end{figure}

\begin{figure}
 \centering
    	\subfloat[Partial metric map.\label{fig:A}]{ \includegraphics[width=0.35\textwidth]{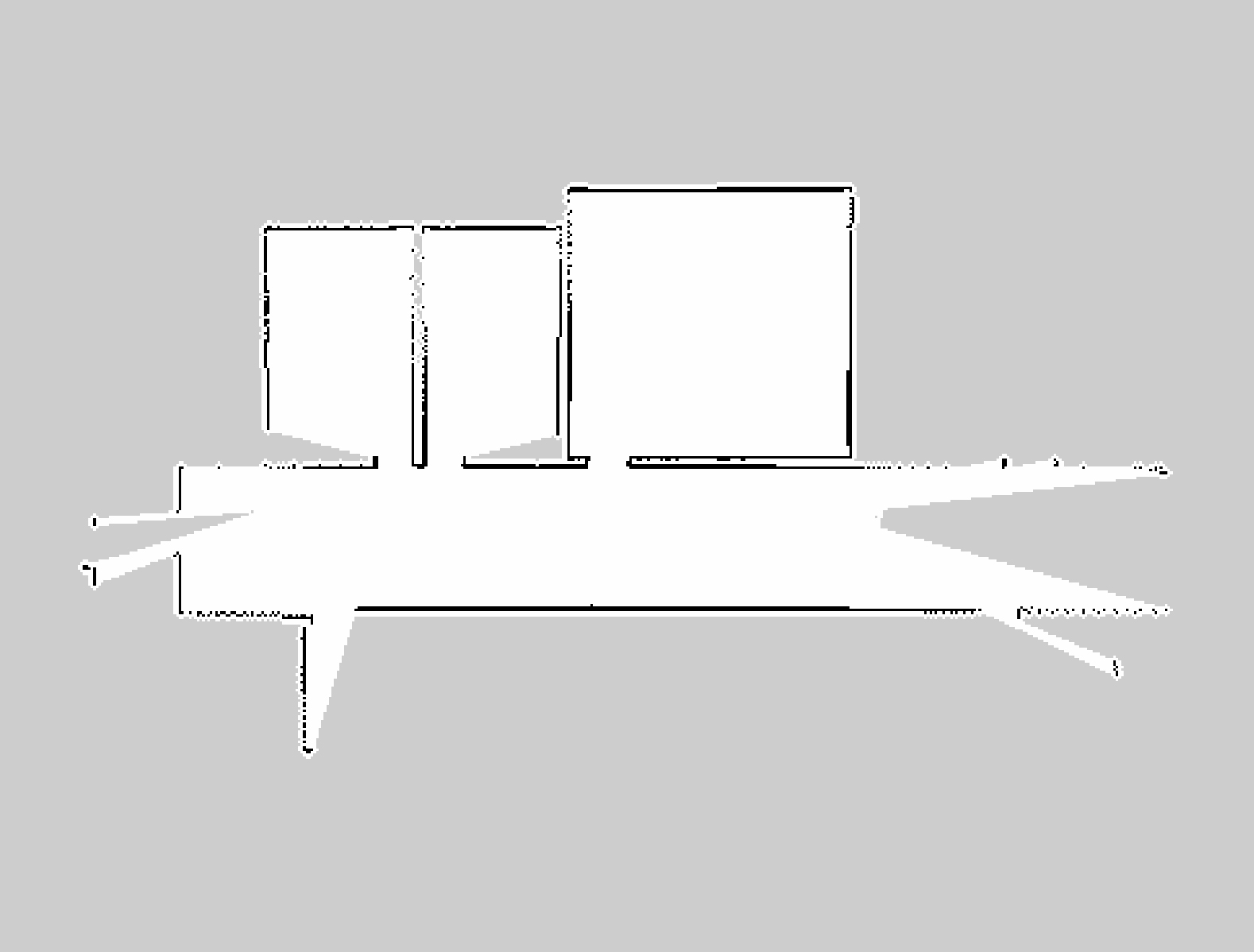}}
    	\subfloat[Predicted layout.\label{fig:B}]{ \includegraphics[width=0.35\textwidth]{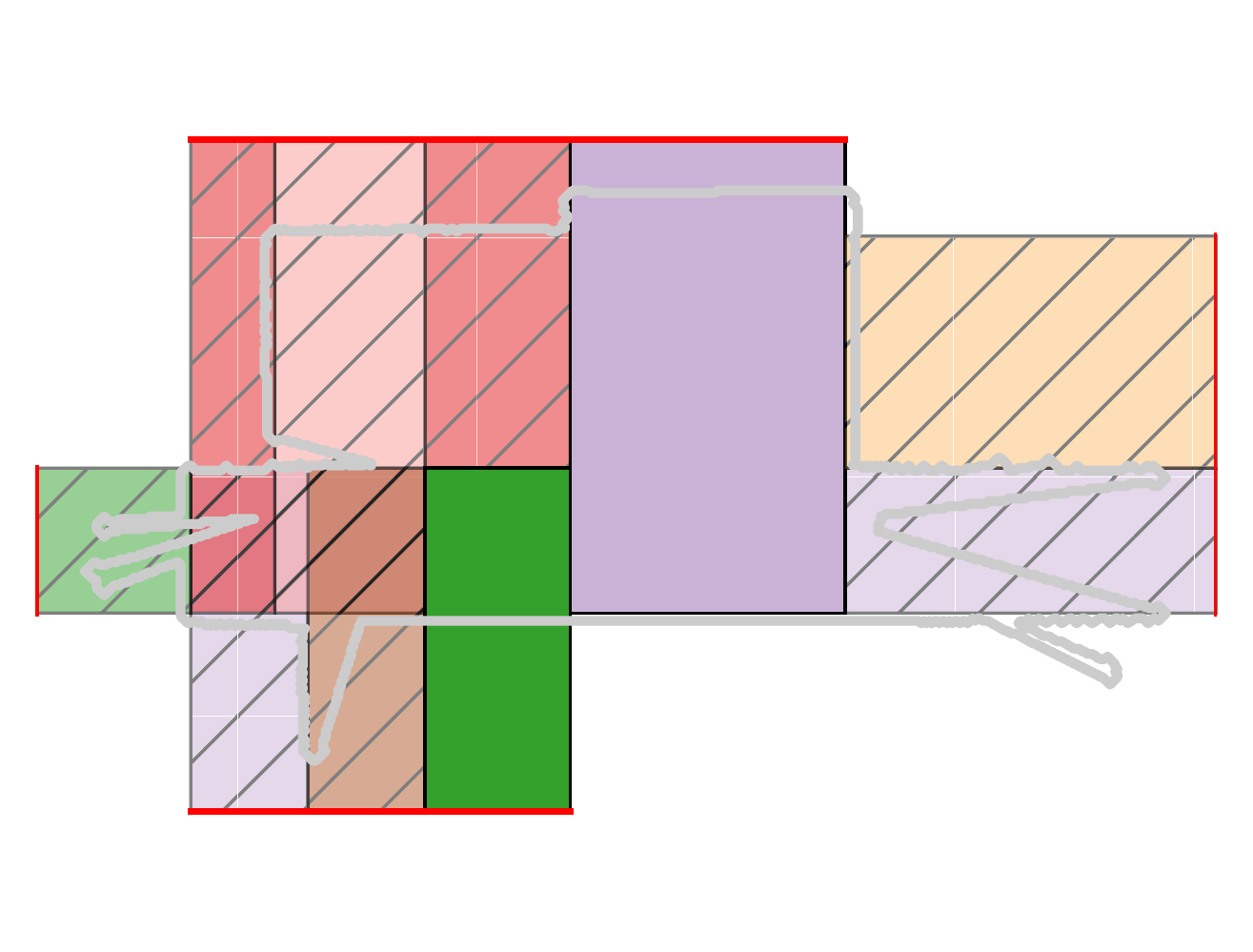}}   	
 \caption{An example of a predicted layout $\mathcal{L}$ obtained with {\it exp} $=30\%$. Despite the fact that only a limited portion of the building is known at this point, the layout $\mathcal{L}$ can provide a good estimate of $I(p)$.\label{fig:EXEP2}} 
\end{figure}

Fig. \ref{fig:EXEP1} shows an example of a reconstructed layout from a grid map with {\it exp} $=80\%$, where most of the partially observed rooms are correctly predicted, despite the fact that relatively large portions of the building are still unobserved. In settings like this one, our proposed method can be a valuable resource in order to select the next best location for robot exploration. In this specific case, the exploration could be stopped because the reconstructed layout correctly represents the actual environment.
Fig.~\ref{fig:EXEP2} shows a partial map obtained during the early stages of an exploration run. It can be noticed how the layout of the rooms is a rough estimate of the correct one, as little information about the shape of the rooms is known at this point. Nevertheless, even in such conditions, our method can provide a reasonable estimate for $I(p)$ for all frontiers.

Further results, including some in more complex environments, are reported in the video at \url{https://youtu.be/OQz-N8xAR6Q}.

\subsection{Discussion}

An interesting, although intuitively expected, result of our experimental analysis is that the availability of a more accurate $I(p)$, using $\mathcal{L}$, produces a speedup in exploration. This fact is remarkably evident at the end of the exploration runs, when $\mathcal{L}$ is more accurate. Indeed, the speedup obtained by our method is not uniformly distributed over the entire exploration process (see Fig.~\ref{fig:PROGR}). Our method performs similarly to the baseline method until, approximatively, the $90\%$ of the total area has been explored. From that moment on, it performs consistently better than the baseline method, eventually resulting in the final gain of Table \ref{tab:results}. On the one hand, the use of an inaccurate prediction about the environment at the early stages of exploration does not jeopardize the gain obtained at the end with an accurate prediction. On the other hand, the use of an inaccurate $\mathcal{L}$ for estimating the information gain produces results similar to those obtained with the mainstream approaches that consider $I(p)$ equal to the footprint of the laser range scanner. When the predicted layout $\mathcal{L}$ becomes enough accurate, the proposed method starts to speed up and, in some cases, the robot does not need to explore anymore, as we discuss in what follows.


We use a very conservative threshold (\SI{1}{\square\meter}) for the ES criterion, with the aim of not stopping the exploration too early when potentially interesting frontiers could still be present, and thus to complete the map with an accurate prediction. In fact, ES is triggered in only $4$ of the $10$ environments we consider and, in those, when the $99\%$ of the area is explored (on average), but provides a remarkable gain in terms of exploration time of $20\%$ (on average). 
However, observing the runs, we note that  $\mathcal{L}$ reliably represents the layout of partially observed rooms when {\it exp} is $80-90\%$. Setting ES to stop exploration at {\it exp}~$=95\%$, we successfully explore our buildings with only minor inaccuracies in $\mathcal{L}$ and a gain of $38\%$ in exploration time. Setting ES to terminate at {\it exp}~$=90\%$ results in an impressive gain of $50\%$ in  exploration time and in missing one room (which is left unexplored because not included in $\mathcal{L}$) for each environment, on average.  
A more aggressive ES criterion that discards candidate locations with low estimated $I(p)$ (according to $\mathcal{L}$) could potentially provide higher gains in exploration time at the risk (related to the accuracy of $\mathcal{L}$) of not exploring some relevant frontiers.

\begin{figure}[t!]
 \centering
 \captionsetup[subfigure]{labelformat=empty}
    	\subfloat{ \includegraphics[width=0.4\textwidth]{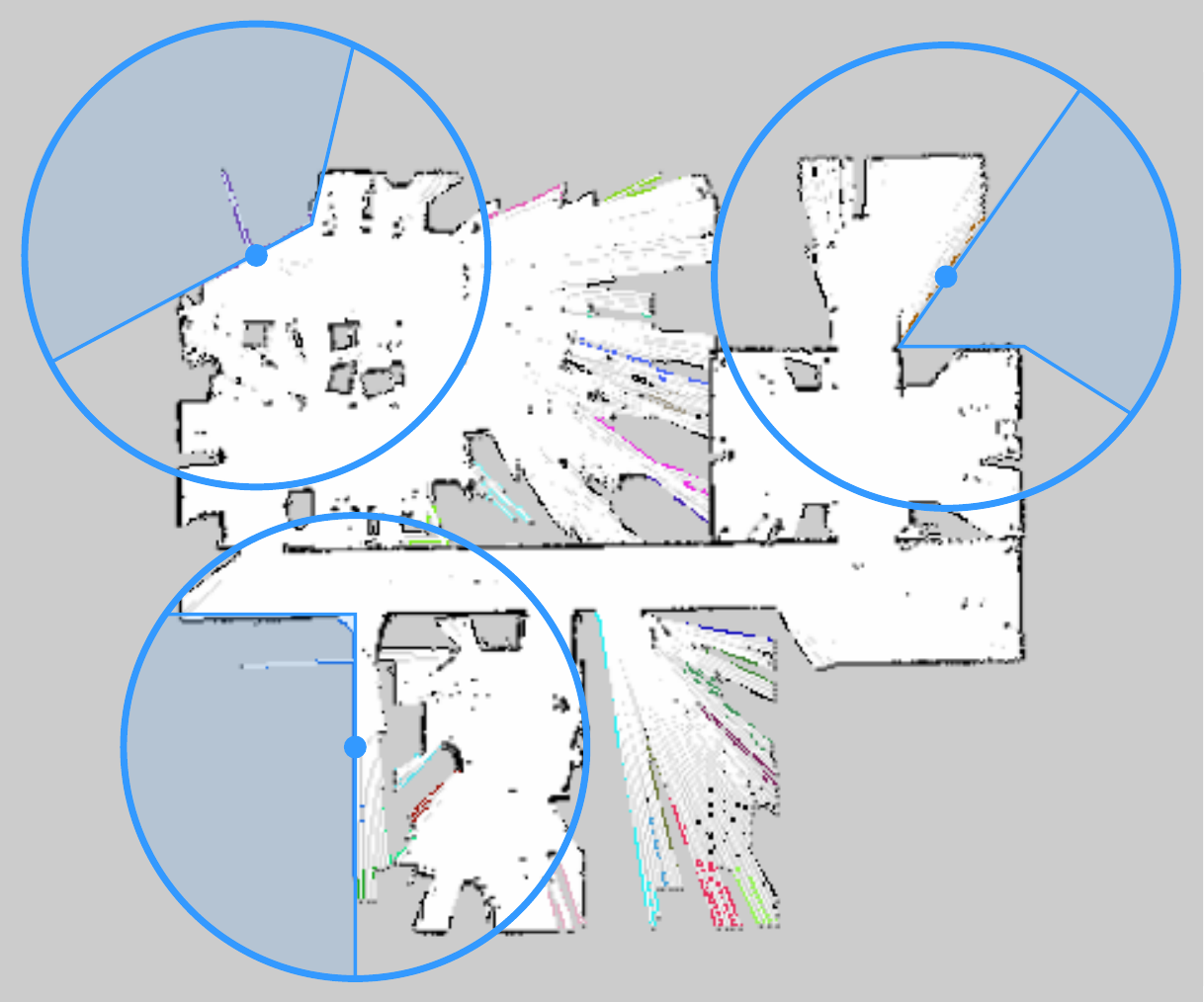}}
    	\subfloat{ \includegraphics[width=0.4\textwidth]{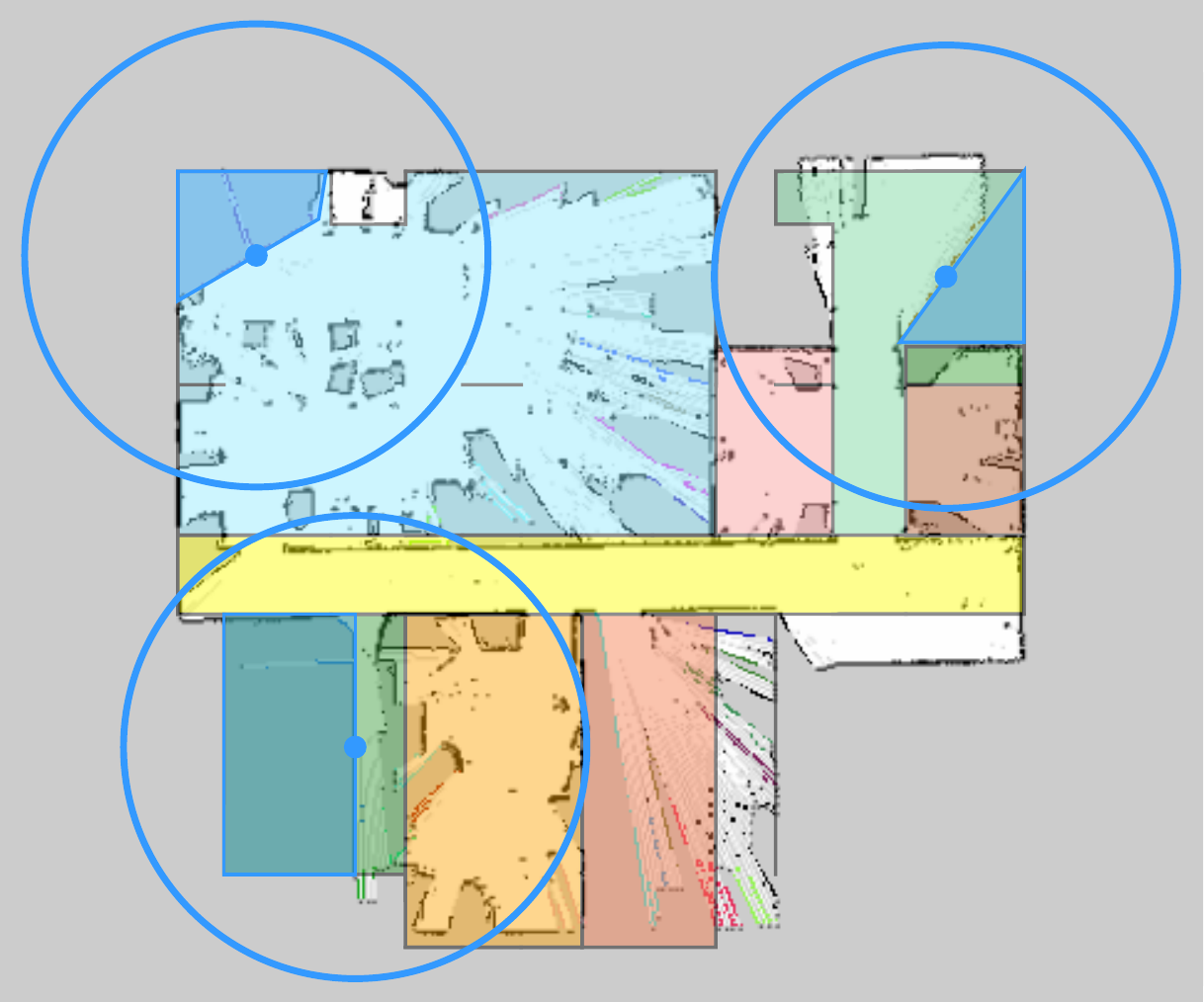}} \\
\caption{$I(p)$ (light blue) of three frontiers in a real-world map with clutter (from \cite{Ruiz-Sarmiento-IJRR-2017}) without (left) and with (right) the use of $\mathcal{L}$. For visualization purposes, we reported a shorter range for the laser range scanner (blue circle) of \SI{2.5}{\meter}. \label{fig:REAL}} 
\end{figure}

Finally, our approach can be reliably applied to partial grid maps acquired in real environments. An example of $I(p)$ computed in a map obtained by a real robot (from \cite{Ruiz-Sarmiento-IJRR-2017}) is shown in Fig. \ref{fig:REAL}. Our method for retrieving the layout filters out the clutter and provides a good estimate of the missing parts of the rooms, for all the three frontiers of the example.

%% file: 05-Con.tex
\section{Conclusions}

In this paper, we have presented a method that shortens the time required by a robot for exploring an initially unknown indoor environment by selecting the next locations according to a predicted layout of the partially observed rooms. 
Experimental results show that our method outperforms state-of-the-art exploration strategies, similar to those of \cite{doi:10.1177/0278364902021010834,Basilico2011}, especially when the predicted layout is accurate. The use of an Early Stopping (ES) criterion, which ends exploration when only uninteresting frontiers are left, could further improve performance.

In addition to devising a more aggressive ES criterion as discussed in the previous section, future work will study a dynamic switch from a classical exploration strategy to our method when the layout $\mathcal{L}$ is enough accurate. Moreover, combining prior knowledge (as in~\cite{AAMAS19,ecmr2019,7390002}) and layout prediction could be investigated, especially in exploring environments for which previous partial maps are available. Finally, experiments with real robots will further assess the improvement provided by our approach to the exploration process.

%% file: arxiv.bbl
\begin{thebibliography}{10}
\providecommand{\url}[1]{\texttt{#1}}
\providecommand{\urlprefix}{URL }
\providecommand{\doi}[1]{https://doi.org/#1}

\bibitem{4543637}
Amigoni, F.: Experimental evaluation of some exploration strategies for mobile
  robots. In: Proc. ICRA. pp. 2818--2823 (2008)

\bibitem{AAMAS19}
Amigoni, F., Fusi, D., Luperto, M.: Exploiting inaccurate a priori knowledge in
  robot exploration (extended abstract). In: Proc. {AAMAS}. pp. 2102--2104
  (2019)

\bibitem{1570708}
Amigoni, F., Gallo, A.: A multi-objective exploration strategy for mobile
  robots. In: Proc. ICRA. pp. 3850--3855 (2005)

\bibitem{ARMENI}
Armeni, I., Sener, O., Zamir, A., Jiang, H., Brilakis, I., Fischer, M.,
  Savarese, S.: {3D} semantic parsing of large-scale indoor spaces. In: Proc.
  {CVPR}. pp. 1534--1543 (2016)

\bibitem{aydemir2012}
Aydemir, A., Jensfelt, P., Folkesson, J.: What can we learn from 38,000 rooms?
  {Reasoning} about unexplored space in indoor environments. In: Proc. {IROS}.
  pp. 4675--4682 (2012)

\bibitem{Basilico2011}
Basilico, N., Amigoni, F.: Exploration strategies based on multi-criteria
  decision making for searching environments in rescue operations. Auton Robot
  \textbf{31}(4),  401--417 (2011)

\bibitem{CaleyHollinger}
Caley, J., Lawrance, N., Hollinger, G.: Deep learning of structured
  environments for robot search. In: Proc. IROS. pp. 3987--3992 (2016)

\bibitem{p-slam}
Chang, J., Lee, G., Lu, Y.H., Hu, C.: {P-SLAM}: Simultaneous localization and
  mapping with environmental-structure prediction. {IEEE T Robot}
  \textbf{23}(2),  281--293 (2007)

\bibitem{doi:10.1177/0278364902021010834}
Gonz{\'a}lez-Ba{\~n}os, H., Latombe, J.: Navigation strategies for exploring
  indoor environments. Int J Robot Res  \textbf{21}(10-11),  829--848 (2002)

\bibitem{gmapping2007tro}
Grisetti, G., Stachniss, C., Burgard, W.: Improved techniques for grid mapping
  with {Rao-Blackwellized} particle filters. {IEEE T Robot}  \textbf{23},
  34--46 (2007)

\bibitem{julia-exploration-survey}
Julia, M., Gil, A., Reinoso, O.: A comparison of path planning strategies for
  autonomous exploration and mapping of unknown environments. Auton Robot
  \textbf{33}(4),  427--444 (2012)

\bibitem{liu2014generalizable}
Liu, Z., von Wichert, G.: A generalizable knowledge framework for semantic
  indoor mapping based on {Markov} logic networks and data driven {MCMC}.
  Future Gener Comp Sy  \textbf{36},  42--56 (2014)

\bibitem{lupertoIAS13}
Luperto, M., Amigoni, F.: Exploiting structural properties of buildings towards
  general semantic mapping systems. In: Proc. IAS-13. pp. 375--387 (2014)

\bibitem{IAS15}
Luperto, M., Amigoni, F.: Extracting structure of buildings using layout
  reconstruction. In: Proc. IAS-15 (2018)

\bibitem{ICRA19}
Luperto, M., Arcerito, V., Amigoni, F.: Predicting the layout of partially
  observed rooms from grid maps. In: Proc. {ICRA}. pp. 6898--6904 (2019).
  \doi{10.1109/ICRA.2019.8793489}

\bibitem{ecmr2019}
Luperto, M., Fusi, D., Borghese, A., Amigoni, F.: Robot exploration using
  knowledge of inaccurate floor plans robot exploration using knowledge of
  inaccurate floor plans. In: Proc. {ECMR} (2019)

\bibitem{luperto2013}
Luperto, M., Quattrini~Li, A., Amigoni, F.: A system for building semantic maps
  of indoor environments exploiting the concept of building typology. In: Proc.
  RoboCup. pp. 504--515 (2013)

\bibitem{luperto2018predicting}
Luperto, M., Amigoni, F.: Predicting the global structure of indoor
  environments: A constructive machine learning approach. Auton Robot  (2018)

\bibitem{mura2014automatic}
Mura, C., Mattausch, O., Villanueva, A., Gobbetti, E., Pajarola, R.: Automatic
  room detection and reconstruction in cluttered indoor environments with
  complex room layouts. Comput Graph  \textbf{44},  20--32 (2014)

\bibitem{neufert2012architects}
Neufert, E., Neufert, P.: Architects' data. John Wiley \& Sons (2012)

\bibitem{7390002}
O{\ss}wald, S., Bennewitz, M., Burgard, W., Stachniss, C.: Speeding-up robot
  exploration by exploiting background information. IEEE RA-L  \textbf{1}(2),
  716--723 (2016)

\bibitem{PereaStrom2017125}
{Perea Str\"{o}m}, D., Bogoslavskyi, I., Stachniss, C.: Robust exploration and
  homing for autonomous robots. Robot Auton Syst  \textbf{90},  125 -- 135
  (2017)

\bibitem{pronobis2012large}
Pronobis, A., Jensfelt, P.: Large-scale semantic mapping and reasoning with
  heterogeneous modalities. In: Proc. {ICRA}. pp. 3515--3522 (2012)

\bibitem{Ruiz-Sarmiento-IJRR-2017}
Ruiz-Sarmiento, J.R., Galindo, C., Gonz{\'{a}}lez-Jim{\'{e}}nez, J.:
  {Robot@Home}, a robotic dataset for semantic mapping of home environments.
  Int J Robot Res  \textbf{36}(2),  131--141 (2017)

\bibitem{LearnedMap}
{Shrestha}, R., {Tian}, F., {Feng}, W., {Tan}, P., {Vaughan}, R.: Learned map
  prediction for enhanced mobile robot exploration. In: Proc. {ICRA}. pp.
  1197--1204 (2019)

\bibitem{Smith2018}
Smith, A., Hollinger, G.: Distributed inference-based multi-robot exploration.
  Auton Robot  \textbf{42}(8),  1651--1668 (2018)

\bibitem{Stachniss2005InformationGE}
Stachniss, C., Grisetti, G., Burgard, W.: Information gain-based exploration
  using {Rao-Blackwellized} particle filters. In: Proc. RSS. pp. 65--72 (2005)

\bibitem{Thrun02roboticmapping:}
Thrun, S.: Robotic mapping: A survey. In: Lakemeyer, G., Nebel, B. (eds.)
  Exploring Artificial Intelligence in the New Millenium, pp. 1--35. Morgan
  Kaufmann (2003)

\bibitem{Tovar2006}
Tovar, B., Munoz, L., Murrieta-Cid, R., Alencastre, M., Monroy, R., Hutchinson,
  S.: Planning exploration strategies for simultaneous localization and
  mapping. Robot Auton Syst  \textbf{54}(4),  314--331 (2006)

\bibitem{1249657}
Tovey, C., Koenig, S.: Improved analysis of greedy mapping. In: Proc. IROS. pp.
  3251--3257 (2003)

\bibitem{613851}
Yamauchi, B.: A frontier-based approach for autonomous exploration. In: Proc.
  CIRA. pp. 146--151 (1997)

\end{thebibliography}
